\documentclass{article} 

\usepackage{iclr2025_conference}
\iclrfinalcopy
\usepackage{times}

\usepackage[T1]{fontenc}
\usepackage{microtype}
\usepackage{graphicx}
\usepackage{tikz}
\usetikzlibrary{patterns}
\usepackage{xfrac}
\usepackage{dsfont}
\usepackage{booktabs} 
\usepackage{minitoc}
\usepackage{floatrow}

\usepackage{tocloft}

\usepackage[super]{nth}
\usepackage{mathrsfs}
\usepackage{nicefrac}
\usepackage{tablefootnote}
\usepackage{enumitem}

\RequirePackage{algorithm}
\RequirePackage{algorithmic}

\usetikzlibrary{bayesnet}
\usetikzlibrary{arrows}

\usepackage{amsthm}
\usepackage{mathtools}
\usepackage[caption=false]{subfig} 

\usepackage[colorlinks=true, linkcolor=blue, citecolor=blue, urlcolor=blue]{hyperref}

\usepackage[capitalize,noabbrev]{cleveref}
\usepackage{autonum}

\usepackage{caption} 

\newcommand{\R}{\mathbb{R}}

\newcommand{\UCB}{\operatorname{UCB}}

\newcommand{\argmax}{\operatornamewithlimits{argmax}}

\newcommand{\Var}{\operatorname{Var}}

\usepackage{amssymb}
\usepackage{amsmath}
\definecolor{royalblue}{rgb}{0.0, 0.0, 0.0}

\theoremstyle{plain}
\newtheorem{theorem}{Theorem}
\newtheorem{proposition}{Proposition}
\newtheorem{lemma}{Lemma}
\newtheorem{corollary}{Corollary}
\theoremstyle{definition}
\newtheorem{definition}{Definition}
\newtheorem{assumption}{Assumption}
\theoremstyle{remark}

\makeatletter
\newcommand\footnoteref[1]{\protected@xdef\@thefnmark{\ref{#1}}\@footnotemark}
\makeatother

\usepackage[textsize=tiny]{todonotes}

\def\rmd{{\rm d}}
\def\SCM{\mathfrak{M}}

\title{Contextual Causal Bayesian Optimisation}

\author{
Vahan Arsenyan \\
CREST, ENSAE, IP Paris, FR \\
\And
Antoine Grosnit \\
Huawei Noah's Ark Lab, UK  
\And
Haitham Bou-Ammar \\
Huawei Noah's Ark Lab, UK \\
\And 
Arnak Dalalyan \\
CREST, ENSAE, IP Paris, FR\\
}

\begin{document}

\maketitle

\begin{abstract}
We introduce a unified framework 
for contextual and causal Bayesian optimisation, 
which aims to design intervention policies 
maximising the expectation of a target variable. 
Our approach leverages both observed contextual 
information and known causal graph structures 
to guide the search. Within this framework, we 
propose a novel algorithm that jointly optimises 
over policies and the sets of variables on which 
these policies are defined. This thereby extends and 
unifies two previously distinct approaches: 
Causal Bayesian Optimisation and Contextual 
Bayesian Optimisation, while also addressing 
their limitations in scenarios that yield 
suboptimal results. We derive worst-case and 
instance-dependent high-probability regret bounds 
for our algorithm. We report experimental results 
across diverse environments, corroborating that 
our approach achieves sublinear regret and reduces 
sample complexity in high-dimensional settings.

\end{abstract}

\doparttoc 
\faketableofcontents 


\raggedbottom
\section{Introduction}

Bayesian optimisation \citep{garnett_bayesoptbook_2023} 
is a powerful approach for optimising expensive black-box 
functions, often formulated as conditional expectations 
of a target variable given that intervenable variables 
are assigned some specific values. 
In settings where intervenable variables exhibit 
causal relationships, specialised Causal Bayesian 
Optimisation (CaBO) algorithms can achieve superior 
performance by strategically focusing on select subsets 
of variables \citep{lee-2018}. The critical challenge 
is to identify these subsets efficiently, since an 
exhaustive search is computationally intractable 
and sample inefficient.

On the one hand, existing CaBO research develops specific 
tools to reduce regret by leveraging the causal graph 
assumed to be known. Consistent with prior work 
\citep{aglietti-2020,zhang2022online}, we restrict
attention to precise interventions \citep[Chapter 3]{pearl-2009} 
and assume the graph is given. Although extensive work on
causal discovery provides methods for inferring such
graphs \citep{Spirtes2000,hoyer-2008,hyttinen-2013,
causal_disc_bo}, this direction lies beyond the
scope of our study. On the other hand, Contextual 
Bayesian Optimisation (CoBO) leverages observable 
context variables but is constrained by a fixed 
policy scope, that is, a fixed tuple of intervenable
and contextual variables. As we demonstrate
in the example in Section \ref{ssec3.3}, this fixed
scope constraint can lead to provably linear regret,
which is suboptimal.
{

\color{royalblue}
Optimising over multiple policy scopes arises in several domains. In complex environments such as video games, fully specifying the state and action spaces is intractable, both computationally and from an optimisation standpoint, so practitioners adopt alternative design choices (\cite{dota}, \cite{starcraft}), each inducing a distinct policy scope. In portfolio optimisation, multiple strategies are proposed, each controlling particular sets of portfolio parameters conditioned on market observables (\cite{cakmak2006portfolio}, \cite{cakmak2020bayesian}). In neural architecture search (\cite{nas}), multiple strategies exist: one may monitor gradient and activation magnitudes and adjust network width and weight initialisation accordingly; another may track test-time performance and the generalisation gap, adjusting learning rate, gradient clipping, and depth. While jointly controlling all architectural parameters and conditioning on all performance metrics might seem optimal, the resulting search space is infeasible, so researchers adopt heuristic, scope-specific strategies. These examples fit naturally within our framework of optimising over multiple scopes.

}

The need for our proposed framework emerges from three 
critical observations discussed above: (1) overcoming 
linear regret in CoBO requires extending beyond fixed 
policy scopes, (2) the exponential growth of possible 
policy scopes with respect to the number of variables 
makes exhaustive search computationally intractable, and
(3) while causal relationships can inform policy scope 
selection, existing CaBO approaches do not incorporate 
contextual variables. In response, we introduce a 
unified framework for contextual and causal Bayesian 
Optimisation (BO).

Our framework, termed CoCa-BO for Contextual-Causal 
Bayesian Optimisation, is designed to reduce the 
search space of policy scopes and guarantee 
convergence to the optimal policy. It addresses the 
challenges of applying the causal acquisition 
function \citep{aglietti-2020} for policy selection 
in contextual settings. We present analytical cases 
where existing methods fail, and show both 
theoretically and experimentally that CoCa-BO 
converges to the optimum. Experimental results 
further demonstrate that CoCa-BO achieves lower 
sample complexity in large-scale systems and 
outperforms existing methods.

Below, we summarise our main contributions:
\vspace*{-8pt}
\begin{itemize}[leftmargin=1.5em] 
\itemsep=0pt
    \item We propose a novel method for 
    context-aware causal Bayesian optimisation 
    that addresses policy selection in 
    context-aware settings with multiple 
    possible policy scopes.
    \item We provide a general, scalable, and 
    practical framework for implementing 
    causality-aware optimisation algorithms, 
    along with a suite of environments for 
    evaluation and benchmarking.
    \item Assuming Gaussian process priors, 
    we establish both worst-case and 
    instance-dependent, high-probability upper 
    bounds on the regret of our method. These
    bounds are valid under mild assumptions on
    the kernels of the Gaussian process priors 
    and highlight the impact of the number of steps on the cumulative regret.
\end{itemize}

\vspace*{-5pt}
\textbf{Notation} Random variables are denoted by 
capital letters like $X$ and their values by 
corresponding lowercase letters like $x$. The domain of 
random variable $X$ is denoted $\mathfrak{X}_X$. 
Sets of variables are denoted by bold capital 
letters and the set of corresponding 
values by bold lowercase letters. Vectors
are always considered as 1-column matrices and for any
vector $\mathbf x$, we denote by $\mathbf x^{\intercal}$
its transpose. Identity matrices are denoted by 
$\mathbf I$ without indicating the dimension, which is
clear from the context. We denote by $\mathbb N^*$ the
set of strictly positive integers, and by $[m]$ 
the set of strictly positive integers smaller than, or equal to, 
$m\in\mathbb N^*$. For $\mathbf{x} \in \mathbb{R}^m$, 
$\|\mathbf{x}\|_1 = \sum_{i=1}^m |x_i|$ and 
$\|\mathbf{x}\|_\infty = \max_{1 \leq i \leq m} 
|x_i|$.

\section{Background}\label{sec:back}

This section aims to provide a background on possibly 
optimal mixed policy scopes and on their importance 
for the contextual decision making setup where the causal 
graph is available.

\vspace*{-8pt}
\paragraph{Preliminaries}

We consider the following general problem: in the absence  
of any intervention, the environment generates a random vector  
$\mathbf{V}$ according to a probability distribution  
denoted by $P_{\mathbf{V}}^0$ on $\mathfrak{X}_{\mathbf V}$. 
We are allowed to perform  
certain interventions (a precise definition will be given  
later) on the components of $\mathbf{V}$, which result in  
modifications of their joint distribution.  
Each set of admissible interventions $\pi$, referred to 
as a  \emph{policy}, induces a new distribution  
$P_{\mathbf{V}}^\pi$ on $\mathfrak{X}_{\mathbf V}$. 
$\Pi$ denotes the set of all admissible policies; it 
will be specified below and may be infinite and 
even uncountable in general.  

We assume that our objective is to identify a  
policy that leads to a nearly maximal expected 
reward. More precisely, for a given reward function 
$r:\mathfrak{X}_{\mathbf V}\to \mathbb R$, our goal is 
to find a policy $\hat{\pi}$ such that 
\begin{align}\label{mu_pi}
    \mu(\hat\pi) := \mathbb{E}_{P^{\hat{\pi}}_{\mathbf{V}}}
    [r(\mathbf{V})],
\end{align}
the expected reward under $\hat\pi$, is close to the 
optimal expected reward  $\sup_{\pi \in \Pi} \mathbb{E
}_{ P^{\pi}_{\mathbf{V}}}[r(\mathbf{V})]$.  
We will focus on the case where $r(\mathbf{V}) = \langle  
\boldsymbol{e}, \mathbf{V} \rangle$, where  
$\boldsymbol{e}$ is an element of the canonical basis.  
This is equivalent to assuming that $Y:=r(\mathbf{V})$ is 
a component  of $\mathbf{V}$.  

To define what a policy is and how the set of admissible  
policies is specified, we follow \citep{lee-2020}.  
Assume that $P_{\mathbf{V}}^0 = P_{\mathbf{V}}^{\SCM}$ is 
given by a structural causal model $\SCM$ 
\citep[Chapter 7]{pearl-2009}.  A structural 
causal model (SCM) $\SCM$ is a quadruple $(\mathbf{V}, 
\mathbf{U}, \mathcal{F}, P_{\mathbf{U}})$, where 
$\mathbf{V}$ and $\mathbf{U}$ are, respectively, a set 
of observed and latent variables. $\mathcal{F}$ is a 
collection of functions such that for each $V \in 
\mathbf{V}$, there exists $f_V \in \mathcal{F}$ that 
determines the value of $V$ based on other variables 
in $\mathbf{V} \cup \mathbf{U}$. That is, for some
$\mathbf{PA}_V \subseteq \mathbf{V}\setminus\{V\}$ 
and $\mathbf{U}_V \subseteq \mathbf{U}$, $V = f_V(\mathbf{PA}_V, \mathbf{U}_V)$. The latent 
variables $\mathbf{U}$ are jointly distributed according to
$P_{\mathbf U}$. Consequently, the SCM $\SCM$ induces a joint 
distribution $P_{\mathbf{V}}^{\SCM}$ over the observed 
variables $\mathbf{V}$. 

The SCM $\SCM$ also defines a directed acyclic graph 
(DAG) $\mathcal{G}$ with two types of nodes. First-type 
nodes represent variables from $\mathbf{V}$, while 
second-type nodes represent  variables from $\mathbf{U}$.  
Only first-type nodes have incoming edges. An edge from 
a node corresponding to a variable $Z$ to a first-type 
node corresponding to $V$ indicates that $Z \in 
\mathbf{PA}_V \cup \mathbf{U}_V$. The policies we 
consider in this paper are defined on mixed policy 
scopes (MPS), which should be compatible 
with the graph $\mathcal{G}$ of the SCM $\SCM$. 
{\color{royalblue} The following definitions and terminology are based on \citep{lee-2020}.
}

\begin{definition}[Mixed policy scope]
    Let $\mathbf{X}^* \subseteq \mathbf{V} \setminus 
    \{Y\}$  and $\mathbf{C}^* \subseteq \mathbf{V} 
    \setminus \{Y\}$ be two sets, termed the set of 
    intervenable variables and the set of contextual 
    variables, respectively. A  \emph{mixed policy scope} 
    $\mathcal{S}$ compatible  with $\mathcal{G}$ is 
    defined as a collection of  pairs $(X; \mathbf{C}_X)$ 
    such that $X \in \mathbf{X}^*$, $\mathbf{C}_X 
    \subseteq\mathbf{C}^*\setminus \{X\}$, and the 
    following compatibility condition is satisfied.  
    The graph $\mathcal{G}_{\mathcal{S}}$ obtained from  
    $\mathcal{G}$ by
    removing edges  
    into $X$ and adding new ones from $\mathbf{C}_X$ to  
    $X$, for every $(X; \mathbf{C}_X) \in \mathcal{S}$,  
    remains acyclic.    
\end{definition}


We denote the set of all MPSs compatible 
with $\mathcal G$ by  $\mathbb{S}[\mathcal{G}]$. 
The notation $\mathbf{X}(\mathcal{S})$ designates  
the set of intervenable variables for an MPS  
$\mathcal{S}$, and $\mathbf{C}(\mathcal{S})$ the  
contextual ones.  


\begin{definition}
    A \emph{mixed policy $\pi$ based on $\mathcal{S}$} is a  
    collection ${\pi} = \{\pi_{X|\mathbf{C}_X}  
    \}_{(X; \mathbf{C}_X)\in \mathcal{S}}$, where  
    $\pi_{X|\mathbf{C}_X}$ is a mapping from 
    $\mathfrak{X}_{\mathbf{C}_X}$ to $\mathfrak{X}_X$.  
\end{definition}
For every MPS $\mathcal{S} \in \mathbb{S}[\mathcal{G}]$,  
the set of all policies based on $\mathcal{S}$ is  
denoted by $\Pi_\mathcal{S}$. We define  
$\Pi_\mathcal{G} = \bigcup_{\mathcal{S} \in \mathbb{S}
[\mathcal{G}]} \Pi_\mathcal{S}$ as the space of all 
policies.  

If $\mathcal{G}$ is the causal graph  
of an SCM $\SCM=(\mathbf{V}, \mathbf{U}, \mathcal{F},P_{\mathbf{U}})$,
every mixed policy $\pi$  
based on $\mathcal{S} \in \mathbb{S}[\mathcal{G}]$  
defines a new SCM  
$\SCM^\pi = (\mathbf{V}, \mathbf{U}, \mathcal{F}', P_{\mathbf{U}})$  
with $\mathcal{F}'$ obtained as follows.  
For each intervenable variable $X\in\mathbf{X} 
(\mathcal{S})$, replace the function $f_X$ from  
$\mathcal{F}$ by the new function $\pi_{X|\mathbf{C}_X}: 
\mathfrak{X}_{\mathbf{C}_X}\to \mathfrak{X}_X$. 
This new SCM $\SCM^\pi$ induces a new distribution
on $\mathbf{V}$, denoted by $P^\pi_{\mathbf{V}}$. 
 
\boxed{\vbox{The agent is given the graph $\mathcal{G}$ 
of an unknown SCM $\SCM$. At each step $t=1,2,\ldots,T$  
\vspace*{-5pt}
\begin{itemize}\itemsep=0pt
    \item[a)] the agent chooses a mixed policy scope 
    $\mathcal S_t\in \mathbb{S}[\mathcal{G}]$ and a mixed 
    policy $\pi_t \in \Pi_{\mathcal{S}_t}$, based on the 
    past observations $\mathcal D^{t-1} = 
    (\mathcal S_\ell,\pi_\ell,\mathbf{c}_\ell, 
    y_\ell)_{1\leqslant \ell\leqslant t-1}$,
    \item[b)] the environment generates $\mathbf v_t$ according to the 
    distribution $P_{\mathbf{V}}^{\pi_t}$,
    \item[c)] the agent observes the components 
    $(\mathbf c_t,y_t)$ of $\mathbf v_t$, corresponding 
    to contextual variables of $\mathcal S_t$ and the 
    reward.
\end{itemize}
\vspace*{-5pt}
The agent aims to minimise the regret
$R_T = T\sup_{\pi\in\Pi_{\mathcal G}} \mu(\pi) 
- \sum_{t=1}^T \mu(\pi_t)$, with $\mu(\pi) = \mathbb E_{P_{\mathbf V}^\pi}[Y]$.
\vspace*{-5pt}
}
}

\paragraph{Possibly-optimal mixed policy scopes}

Our work focuses on a special set of MPSs that are possibly 
optimal under an SCM compatible with $\mathcal{G}$ 
\citep{lee-2020}. Let $\mu^*_{\mathcal S} = \sup_{\pi\in 
\Pi_{\mathcal S}} \mu(\pi)$ be the largest expected reward
corresponding to the MPS $\mathcal S$.

\begin{definition}[Possibly-Optimal MPS (POMPS)]
    Given $(\mathcal{G}, Y, \mathbf{X}^*, \mathbf{C}^*)$, 
    we say that $\mathcal{S}\in\mathbb{S}[\mathcal G]$ is a 
    possibly-optimal MPS if there exists an SCM $\SCM$ 
    compatible with the graph $\mathcal{G}$ such that 
    $\mu^*_{\mathcal{S}} > \mu^*_{\mathcal{S'}}$ for 
    any $\mathcal{S'} \in\mathbb{S}[\mathcal G]$ 
    different from $\mathcal S$\footnote{$\mathcal{S}$ must additionally be non-redundant under optimality. Please see Definition 4 of \cite{lee-2020}.}.

\end{definition}
{
\color{royalblue}
We denote by $\mathbb{S}^*[\mathcal{G}]$ the set of all POMPSs compatible with $\mathcal{G}$. The policy considered in this work selects $\mathcal{S}_t$ from $\mathbb{S}^*[\mathcal{G}]$. While further restricting the candidate set could reduce computational cost, it would risk linear regret: for any $\mathcal{S} \in \mathbb{S}^*[\mathcal{G}]$, there exists an SCM $\SCM$ compatible with $\mathcal{G}$ for which $\mathcal{S}$ is optimal. Thus, given only $\mathcal{G}$, the agent cannot exclude any POMPS from the set of candidate scopes without potentially discarding the optimal policy and incurring linear regret.
}

\vspace*{-8pt}
\paragraph{Example of CoBO/CaBO's failure modes}\label{ssec3.3}

Let us illustrate through an example the 
role of incorporating both causal and contextual 
information. Specifically, we demonstrate that 
neglecting either component can lead to regret that 
is not sublinear in the number of steps~$T$. Consider 
the example in \Cref{Fig:1}, where the context 
variable $C$ influences both the intervenable 
variable $X_2$ and the outcome $Y$. The causal graph
$\mathcal{G}$ also includes the intervenable 
variable $X_1$, which affects $X_2$.  

\begin{figure}[h]
    \centering
    \vspace*{-15pt}
    \subfloat[Graph $\mathcal{G}$]{%
        \label{Fig1:a}%
        \begin{minipage}{0.4\linewidth}
            \centering
            \begin{tikzpicture}[every text node part/.style={align=center}]
              \node[draw,circle, minimum size=.8cm] (X1) at (0, 1.5) {$X_1$};
              \node[draw,circle, minimum size=.8cm] (X2) at (1.5, 1.5) {$X_2$};
              \node[draw,circle,pattern=dots, minimum size=.8cm] (Y) at (3, 1.5) {$Y$};
              \node[draw,circle,fill=gray!30, minimum size=.8cm] (C) at (1.5, 3) {$C$};

              \node (U1) at (0, 3) {$U_1$};
              \node (U2) at (3, 3) {$U_2$};
              \draw[>=triangle 45, ->] (X1) -- (X2);
              \draw[>=triangle 45, ->] (X2) -- (Y);
              \draw[>=triangle 45, ->] (C) -- (X2);
              \draw[>=triangle 45, ->] (C) -- (Y);
              \draw[dashed, ->] (U1) -- (C);
              \draw[dashed, ->] (U1) -- (X1);
              \draw[dashed, ->] (U2) -- (X2);
              \draw[dashed, ->] (U2) -- (Y);
            \end{tikzpicture}
        \end{minipage}%
    }
    \subfloat[Structural causal model]{%
        \label{Fig1:b}%
        \begin{minipage}{0.4\linewidth}
            $$
                \begin{matrix}
                &U_1, U_2 \sim \operatorname{Uniform}([-1, 1]) \\[5pt]
                &X_1 = U_1, \quad C=U_1 \\[5pt]
                &X_2 = U_2 \operatorname{exp}\{ -(X_1 + C)^2\} \\[5pt]
                &Y = U_2 X_2 + C
                \end{matrix}
            $$
        \end{minipage}%
    }

    \caption{An SCM and its DAG $\mathcal G$ for 
    which CoBO and CaBO fail 
    to converge to the optimum.}%
    \label{Fig:1}%
    \vspace*{-10pt}
\end{figure}

First, we observe that there exists a policy 
achieving an expected value of $Y$ equal to 
$\sfrac{1}{3}$. Specifically, consider the mixed policy 
$\pi^{(1)}$ defined on the MPS $\mathcal{S}_1 = 
\{(X_1; C)\}$ by $\pi^{(1)}_{X_1|C}(c) = -c$.
Under this policy, we have $\mathbb{E}_{\pi^{(1)}}[Y] 
= \mathbb{E}_{\pi^{(1)}}[U_2^2] \cdot 
\mathbb{E}_{\pi^{(1)}}\bigl[\smash{e^{-(X_1 + C)^2}}\bigr] + 
\mathbb{E}_{\pi^{(1)}}[ C] = \mathbb{E}[U_2^2] = 
\nicefrac{1}{3}$.

If we ignore the contextual information and rely 
solely on the causal relationships, at each step the 
choice reduces to one of three options: no 
intervention, intervention on $X_1$, or intervention 
on $X_2$. Intervening simultaneously on both $X_1$ 
and $X_2$ is excluded because, according to the graph 
$\mathcal{G}$ depicted above, intervening on $X_2$ 
renders the value of $X_1$ irrelevant to $Y$. In the 
case of no intervention, we have $\mathbb{E}[Y] = 
\sfrac{1}{3} \, \mathbb{E}\bigl[e^{-4U_1^2}\bigr] 
\leqslant \sfrac{1}{6}$. If we intervene on $X_2$, it 
becomes independent of $U_2$ and the expected value of 
$Y$ reduces to zero. Finally, intervening on $X_1$ by 
setting it to a fixed value $x$ yields
$\mathbb{E}_{\operatorname{do}(X_1 = x)}[Y] = \sfrac{1}{3} 
\, \mathbb{E}\bigl[e^{-(x + U_1)^2}\bigr] \leqslant 
\sfrac{1}{3} \, \mathbb{E}\bigl[e^{-U_1^2}\bigr] < 
\sfrac{1}{4}$. Therefore, for every policy $\pi$ that 
does not incorporate the context, the expected reward 
is strictly less than $\sfrac{1}{4}$, and consequently, 
the cumulative regret of any such algorithm is at 
least $T/12$.

Finally, suppose we ignore the causal structure and 
rely solely on the contextual information. Then, the 
decision reduces to selecting a function $\pi^{(2)} : 
c \mapsto (x_1, x_2)$ that specifies the values of 
$(X_1,X_2)$ given $C = c$. Under such a policy,  
$\mathbb{E}_{\pi^{(2)}}[Y] = \mathbb{E}_{\pi^{(2)}} 
[U_2 X_2 + C] = \mathbb{E}\bigl[U_2 \, \pi^{(2)}_2(C) 
+ C\bigr]$. Note that $\mathbb{E}[U_2] = \mathbb{E}[C] = 0$,
and $\pi^{(2)}_2(C)$ being a deterministic function of $C$, 
is independent of $U_2$. Therefore, $\mathbb{E}_{\pi^{(2)}} 
[Y] = 0$. This shows that, for any context-only 
policies used across the $T$ steps, the expected reward 
is zero. Thus, neglecting the causal structure
leads to a regret of at least $T/3$.

\raggedbottom

\section{Proposed algorithm}\label{prop}

In this section, we provide high-level description of the identification of 
intervenable scopes and the selection of optimal 
policies. We then formalise the contextual BO strategy used within each scope. Finally, 
we discuss why CaBO’s POMPSs selection strategy is inadequate in contextual settings and motivate our alternative approach.

\vspace*{-8pt}
\paragraph{High-level description}
{
\color{blue}
\begin{figure}[h]
    \centering
    \includegraphics[width=1\linewidth]{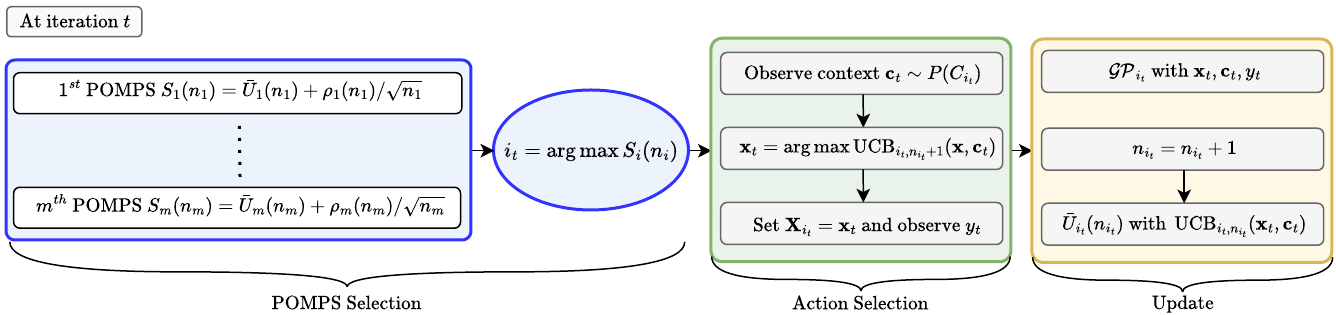}
    \caption{\textcolor{royalblue}{At each iteration $t$, we select a POMPS using the running average of the UCB evaluated at the points $\mathbf{x}_t$ and $\mathbf{c}_t$, plus an exploration term $\rho_i(n_i)/\sqrt{n_i}$, where $n_i$ is the number of times POMPS $i$ has been chosen so far. We then implement the intervention $\mathbf{X}_{i_t}$, observe the target $y_t$ under this intervention, and update the parameters of the algorithm. For more details please see \Cref{alg:coca_bo}.}}
    \label{fig:schematics}
\end{figure}
}

Our method, outlined in Alg.~\ref{alg:CoCa-BO}, 
has 3 main components. The first component is the 
$\operatorname{POMPS}$ function, which computes the set 
$\mathbb{S}^* = \mathbb{S}^*[\mathcal{G}]$ 
\citep{lee-2020} for a given causal graph $\mathcal{G}$. 
This computation involves a brute-force search 
over MPSs, with computational complexity scaling 
exponentially in $|\mathbf{V}|$. However, this step 
is highly parallelisable, a feature we exploit in 
our implementation. Our definition of the contextual
variables as $\mathbf{C}^* = \mathbf{V} \setminus (\mathbf{X}^* \cup \{Y\})$ suffices to ensure that 
optimal POMPSs are not missed 
\citep[Prop.~2]{lee-2020}. A visual overview is provided in \Cref{fig:schematics}.

The method proceeds as follows: at each step $t$, based on 
historical data $\mathcal{D}^{t-1} := \{(\mathcal{S}_i, \pi_i, 
\mathbf{c}_i, y_i)\}_{i \in [t-1]}$, it first selects a POMPS 
$\mathcal{S}_t$ from $\mathbb{S}^*$, then chooses a policy 
$\pi_t \in \Pi_{\mathcal{S}_t}$. This amounts to choosing an 
interventional value for $\mathbf{X}_{\mathcal{S}_t}$ after 
observing the context $\mathbf{c}_t$ associated with the 
variables $\mathbf{C}_{\mathcal{S}_t}$. Consequently, we may 
equivalently write the historical data as $\mathcal{D}^{t-1} = 
\{(\mathcal{S}_i, \mathbf{c}_i, \mathbf{x}_i, y_i)\}_{i \in [t-1]}$. 
The second component is the selection of a POMPS from the finite 
set $\mathbb{S}^*$ using $\operatorname{MAB}$. Each POMPS serves as 
an arm, and the algorithm aims to find the one with the largest 
mean reward $\mu^*_{\mathcal{S}}$. Formally, we may view 
$\operatorname{MAB}$ as a mapping $A: \mathbb{S}^* 
\to \mathbb{R}$. At each step $t$, it selects a POMPS 
$\mathcal{S}_t \in \operatorname{argmax}_{\mathcal{S} \in 
\mathbb{S}^*} A(\mathcal{S})$, and after observing 
$(\mathbf{c}_t, \mathbf{x}_t, y_t)$, it updates the value 
$A$ at $\mathcal{S}_t$ while keeping all others unchanged. 
Details of this update step are provided in \Cref{alg:coca_bo}. 
The third component of our method is Bayesian Optimisation 
routine with Gaussian process surrogate, 
used to choose the intervention value of 
$\mathbf{X}_{\mathcal{S}_t}$ once the POMPS $\mathcal{S}_t$ is 
selected and the context $\mathbf{c}_t$ is observed. This 
component is described in more detail later in this section. 
Finally, the selected BO optimiser is updated at each 
iteration using the newly observed values. These updates 
involve Gaussian process posterior updates, which require 
inverting a $T \times T$ matrix, and thus have computational
complexity $\mathcal{O}(T^3)$.

\begin{algorithm}[t]
   \caption{Contextual Causal BO (CoCa-BO)}
   \label{alg:CoCa-BO}
\begin{algorithmic}
   \STATE {\bfseries Input:} causal graph $\mathcal{G}$ on the 
   nodes $\mathbf{V}$, the target $Y\in\mathbf{V}$, the set of 
   intervenable variables $\mathbf{X}^*\subset \mathbf{V}$, 
   domain of variables $\mathfrak{X}_\mathbf{V}$, number of 
   iterations $T$. We consider
   $\mathbf{C}^* = \mathbf{V}\setminus (\mathbf{X}^*\cup \{Y\})$. 
   \STATE Compute $\mathbb{S}^*=\operatorname{POMPS}(\mathcal{G}, 
   \mathbf{X}^*) $ and create  $A=\operatorname{MAB(\mathbb{S}^*)}$
   
   \STATE \textbf{for} $\mathcal{S}\in\mathbb{S}^*$ \textbf{do} 
Create $\textnormal{Opt}_\mathcal{S}=\operatorname{BO}
(\mathbf{X}(\mathcal{S}), \mathbf{C}(\mathcal{S}), \mathfrak{X}_\mathcal{S})$
   \FOR{$t=1$ {\bfseries to} $T$}
   \STATE Select a POMPS: $\mathcal{S}=A.\textnormal{suggest}()$
   \STATE Implement $\pi_{\mathcal{S}}$ provided by 
   $\textnormal{Opt}_{\mathcal{S}}$, 
   \STATE Observe $y, \mathbf{x_\mathcal{S}},\mathbf{c_{\mathcal{S}}}$ 
   drawn from $P_{\mathbf V}^{\pi_{\mathcal{S}}}$
   \STATE $\textnormal{Opt}_{\mathcal{S}}.\textnormal{update}(y, \mathbf{x_\mathcal{S}},\mathbf{c_{\mathcal{S}}})$; $A.\textnormal{update}(y, \mathcal{S}, \textnormal{Opt}_{\mathcal{S}})$
   
   \ENDFOR
\end{algorithmic}
\end{algorithm}

\vspace*{-8pt}
\paragraph{Contextual Bayesian optimisation policy}

For a given scope $\mathcal{S}$ selected at step $t$, the 
contextual Bayesian optimisation assumes that 
$y_t = g(\mathbf{x}_t,\mathbf{C}_t) + \sigma_{\text{n}} \xi_t$, 
where $g$ is a random Gaussian process defined on 
$\mathfrak{X}_{\mathbf{X}_{\mathcal{S}}}\times \mathfrak{X}_{\mathbf{C}_{\mathcal{S}}}$, $\mathbf{C}_t$ is a 
random vector drawn from $P^0_{\mathbf{C}_{\mathcal{S}}}$, and 
$\xi_t$ is a standard Gaussian random variable. All three 
quantities $(g, \mathbf{C}_t, \xi_t)$ are assumed independent. 
To provide more details, we define the set $[t]_\mathcal{S} 
= \{i \in [t] : 
\mathcal{S}_i = \mathcal{S} \}$ and denote by 
$\mathcal{D}_\mathcal{S}^t = \{(\mathbf{x}_{\mathcal{S}}^i, 
\mathbf{c}_{\mathcal{S}}^i, y^i)\}_{i \in [t]_\mathcal{S}}$ the 
subsample of $\mathcal{D}^t$ corresponding to those steps $i$ 
where the scope $\mathcal{S}$ was selected. The Gaussian process 
$g$ has zero mean and covariance function $k:
(\mathfrak{X}_{\mathbf{X}_{\mathcal{S}}} \times 
\mathfrak{X}_{\mathbf{C}_{\mathcal{S}}})^2 \to \mathbb{R}$. 
Given the dataset $\mathcal{D}_\mathcal{S}^t$, the posterior 
distribution of the output for a test point $(\mathbf{x}, 
\mathbf{c})$ is a Gaussian 
$\mathcal{N}(\mu_\text{post}(\mathbf{x}, \mathbf{c}), 
\sigma^2_\text{post}(\mathbf{x}, \mathbf{c}))$, where
\begin{align}
\mu_\text{post}(\mathbf{x}, \mathbf{c}) &= 
\boldsymbol{k}_{[t]_\mathcal{S}}^{\intercal}(\mathbf{x}, 
\mathbf{c}) \left( \boldsymbol{K}_{[t]_\mathcal{S}} + 
\sigma_{\text{n}}^2 \mathbf{I} \right)^{-1} 
\mathbf{y}_{[t]_\mathcal{S}} \\
\sigma^2_\text{post}(\mathbf{x}, \mathbf{c}) &= 
k((\mathbf{x}, \mathbf{c}), (\mathbf{x}, \mathbf{c})) - 
\boldsymbol{k}_{[t]_\mathcal{S}}^{\intercal}(\mathbf{x}, 
\mathbf{c}) \left( \boldsymbol{K}_{[t]_\mathcal{S}} + 
\sigma_{\text{n}}^2 \mathbf{I} \right)^{-1} 
\boldsymbol{k}_{[t]_\mathcal{S}}(\mathbf{x}, \mathbf{c}),
\label{sigma_post}
\end{align}
with $\mathbf{I}$ the identity matrix of appropriate size, 
$\mathbf{y}_{[t]_\mathcal{S}}$ the vector of observed targets in 
$\mathcal{D}_\mathcal{S}^t$, $\boldsymbol{K}_{[t]_\mathcal{S}}$ 
the kernel matrix of all $(\mathbf{x}, \mathbf{c})$ pairs in 
$\mathcal{D}_\mathcal{S}^t$, and 
$\boldsymbol{k}_{[t]_\mathcal{S}}(\mathbf{x}, \mathbf{c})$ the 
vector of covariances between the test point and each training 
point. The noise level $\sigma_{\text{n}}$ is a tunable parameter.

In our method, we deploy HEBO \citep{hebo} for each POMPS, as it 
is a state-of-the-art BO algorithm \citep{turner-21} capable of 
optimising in contextual settings and handling both continuous 
and discrete variables. HEBO's Gaussian process model includes 
learnable transformations of both input and output, which makes 
it robust to heteroscedasticity and non-stationarity. All methods 
described in \Cref{sec:experiments} use HEBO as the BO subroutine.

\vspace*{-8pt}
\paragraph{POMPS selection}

Our algorithm performs POMPS selection using historical values of the acquisition function combined with a margin (see \Cref{app:main}).
In contrast, CaBO selects the next intervenable set based on 
the values of the acquisition functions associated with each 
GP. This strategy is motivated by the 
intuition that the optimal intervenable set, together with its
optimal interventional values, should yield the highest reward. 
Accordingly, acquisition functions are expected to reflect this 
pattern, particularly in regions of low uncertainty. 
However, this approach becomes unsuitable in contextual settings, 
where the optimal values of acquisition functions inherently 
depend on the observed context. This dependence introduces a 
fundamental issue: the comparison is performed between acquisition 
functions evaluated at different contexts, which undermines the 
validity of such comparisons. 
Such comparisons may lead the optimisation 
process to switch arbitrarily between scopes based on the 
observed context, potentially causing instability or divergence.

{
\color{royalblue}

Our selection criterion $\bar U_{i_t}$, defined in \Cref{alg:coca_bo}, is the running average of the $\UCB$ values over the evaluation points for each POMPS instance. As shown in \Cref{lem:bound_on_smp_mean}, this quantity simultaneously upper and lower bounds the empirical mean of the target under the optimal policy, up to additive terms. These terms decay sufficiently fast to ensure that suboptimal POMPS are selected only rarely, as formalized in \Cref{lem:sub_opt_play}.

}

\section{Regret Bound for CoCa-BO}\label{main_th_sec}


In this section, we prove that the CoCa-BO algorithm 
outlined in \Cref{alg:coca_bo} achieves sublinear regret. 
For this purpose, we place ourselves in a more general 
setting, which we refer to as multi-function BO. 
We assume that we are given an integer $m \in \mathbb{N}^*$, 
together with two sequences of sets 
$(\mathfrak{X}_{\mathbf{X}(i)})_{i \in [m]}$ and 
$(\mathfrak{X}_{\mathbf{C}(i)})_{i \in [m]}$. 
At each step $t \in \mathbb{N}^*$, we first select 
an index $i_t \in [m]$ and then observe a random variable 
$\mathbf{C}_t$ drawn from $\mathfrak{X}_{\mathbf{C}(i_t)}$. 
Based on $\mathbf{C}_t$ and the past observations, 
we choose $\mathbf{X}_t$ from $\mathfrak{X}_{\mathbf{X}(i_t)}$ 
and subsequently observe the reward $Y_t$. In this 
formulation, the indices $i \in [m]$, henceforth 
called \emph{arms}, correspond to POMPSs; the variables 
$\mathbf{C}_t$ correspond to \emph{contexts}; and the 
variables $\mathbf{X}_t$, henceforth called \emph{actions}, 
correspond to interventional values.

We assume that for some random processes $f_1,\ldots,f_m$
and for some unknown context distributions $P_1,\ldots,P_m$, 
the reward takes the form $Y_{i_t}=f_{i_t}(\mathbf{X}_{t},
\mathbf C_t)+\epsilon_t$, where 
conditionally to $i_t,f_1,\ldots,f_m$, $\mathbf C_t$ is 
drawn from $P_{i_t}$, and conditionally to $i_t,f_1,\ldots,
f_m$ and $\mathbf C_t$, $\epsilon_t{\sim}\mathcal{N}(0,
\sigma^2)$. A more detailed description of the setting can
be found in \Cref{ssec:setting}. 

\def\dprimei{d'\!\!{}_i}
\begin{assumption}[Convex-compact domains]\label{ass:1}
    For every $i\in[m]$, $\mathfrak{X}_{\mathbf X(i)}$ 
    and $\mathfrak{X}_{\mathbf C(i)}$ are convex and
    compact. Furthermore, there exist constants $r_i, 
    r_i'>0$, $d_i,\dprimei\in\mathbb N$ such that 
    $\mathfrak{X}_{\mathbf X(i)}\subset [0,r_i]^{d_i}$ 
    and $\mathfrak{X}_{\mathbf C(i)} \subset [0,r'_i]^{
    \dprimei}$ for every $i\in[m]$. We set $\bar r_i = 
    r_i \vee r_i'$ and $\bar d_i = d_i\vee \dprimei $.
\end{assumption}

\begin{assumption}[Smooth and bounded kernels]\label{ass:2}
    Processes $f_i$ are independent, each $f_i$ is Gaussian
    with zero mean and covariance kernel $k_i$. The covariance kernels $k_i$ are bounded, \textit{i.e.}, 
    for some $\kappa_1,\ldots,\kappa_m>0$, we have
    $\max_{\mathbf x,\mathbf c} k_i\big((\mathbf x,\mathbf c);
    (\mathbf x,\mathbf c)\big)\le \kappa_i$. Furthermore, there
    are constants $\psi,\varphi>0$ such that, for any 
    $\delta\in(0,1)$ and for any $i\in[m]$, with probability 
    at least $1-\delta$
    \begin{align}\label{eq:maxF}
        \smash{\sup_{\mathbf x,\mathbf c}}\|\nabla f_i(\mathbf 
        x,\mathbf c)\|_\infty \le \varphi \sqrt{
        \log(\smash{\bar d_i}\psi/\delta)}.
    \end{align}
    \vspace*{-10pt}
\end{assumption}
Note that these assumptions are not very restrictive 
compared to those usually imposed in Bayesian optimisation. 
In particular, it follows from \citep{Ghosal_Roy} that 
if the kernels $k_i$ admit fourth-order mixed partial 
derivatives, then the path smoothness condition 
\eqref{eq:maxF} is satisfied.

For brevity, we defer the precise definitions of the 
subroutines MAB-UCB and HEBO from \Cref{alg:CoCa-BO} 
to \Cref{app:main}. We only mention here that, in the 
Bayesian optimisation subroutine, an upper confidence 
bound is constructed from the posterior means 
$\mu^{(i)}_{t-1}$ and standard deviations 
$\sigma^{(i)}_{t-1}$ of the $i$-th GP at time $t$, 
together with a sequence of tuning parameters 
$(\beta_i(n))_{i \in [m],\, n \in \mathbb{N}^*}$. 
As for the arm-selection subroutine MAB-UCB, it 
chooses the arm with the largest average UCB over 
past observations, plus a penalty term involving the 
sequence of tuning parameters $(\rho_i(n))_{i \in 
[m],\, n \in \mathbb{N}^*}$. To ease that statement
of the main theorem, we set
\begin{align}
    \beta_T = \max_{i\in[m]}\max_{n\le T} 
    \beta_i(n)\qquad \text{and }\qquad
    \rho_T = \max_{i\in[m]}\max_{n\le T} 
    \rho_i(n).
\end{align}
Let $\mu_i = \mathbb{E}[f_i^*(\mathbf{C}_{i,1}) 
\,|\, f_i\,] = \int \max_{\mathbf x} f_i(\mathbf x,
\mathbf c)\, P_i(\mathrm d\mathbf c)$ and $I^* = 
\argmax_{i\in[m]} \mu_i$. Clearly, $\mu_i$ depends 
on $f_i$ and $I^*$ depends on $\boldsymbol{f} = 
(f_1,\ldots,f_m)$, but for simplicity of notation 
we do not make this dependence explicit. We also 
write $\mu^* = \max_{i\in[m]} \mu_i$. The cumulative 
regret is then
\begin{align}\label{eq:reg_def0}
    R_T &= 
    \sum_{t=1}^T
    \big\{\mu^* - 
    \mathbb E\big[f_{i_t}(\mathbf{X}_{t}, \mathbf{C}_t)
    \mid \boldsymbol{f}, \mathcal D^{t-1}\big]\big\}.
\end{align}
In words, the regret is the difference between 
the context-averaged reward of an oracle strategy 
(which always selects an arm with maximal 
context-averaged reward and plays a context-optimal 
action) and the context-averaged reward of our 
algorithm. This notion of regret, specific to 
multi-function BO, differs from those considered 
in the MAB literature \citep{Lattimore_Szepesvári_2020} 
and the BO literature \citep{srinvas_kauss}. 
In MAB, the reward of a chosen arm is constant across 
rounds, whereas in our setting, the expected reward 
evolves as the estimated maxima of $f_{i_t}$ are 
progressively refined. On the other hand, unlike in BO, 
our regret compares the context-averaged maximum of 
$f_{i_t}$ to $\mu^*$, which may correspond to the 
optimal value of another function $f_j$. 
These distinctions introduce additional challenges 
in establishing a bound for regret $R_T$ of the 
procedure defined in Algorithm~\ref{alg:CoCa-BO}.

A key quantity characterising the regret of the 
algorithms commonly used in BO with Gaussian process 
prior is the maximum information gain 
\citep{srinvas_kauss}. Denoted by $\gamma_i(n)$, 
the maximum information gain of the $i^{th}$ GP 
$f_i$ over horizon $n$, is defined by
\begin{align}
    \gamma_i(n) = \max_{A:|A| = n} 
    \frac12 \log\det(\mathbf I + \sigma^{-2} 
    \boldsymbol{K}_{i,A}),
\end{align}
where the maximum is over all sets $A\subset 
\mathfrak{X}_{\mathbf X(i)}\times \mathfrak{X}_{
\mathbf C(i)}$ of cardinality $n$ and 
$\boldsymbol{K}_{i,A}$ is the $n\times n$ matrix
with the general term $K_i(\mathbf z,\mathbf z')$, 
$\mathbf z,\mathbf z'\in \mathfrak{X}_{\mathbf X(i)}
\times \mathfrak{X}_{\mathbf C(i)}$. We refer the 
reader to \citep{inf_gain_bounds} for results 
on bounding the maximum information gain 
in terms of $n$ and the kernel parameters. Based 
on $\gamma_i(n)$, we define the  
worst-case-arm and arm-averaged information gains by
\begin{align}
    \gamma_T = \max_{i\in [m]} \gamma_i(T),
    \qquad\text{and}\qquad 
    \bar\gamma_T = \max_{\mathbf n:\sum_{i} n_i = T}
    \frac1m\sum_{i=1}^m \gamma_i(n_i),
\end{align}
where the last maximum is over all vectors 
$\mathbf n=(n_1,\ldots,n_m)\in\mathbb N^m$ summing 
to $T$.
\begin{theorem}\label{th:1}
    Let $\delta\in(0,1/2)$,  $\boldsymbol{\Delta} 
    = (\Delta_1,\ldots,\Delta_m)$ be the vector of 
    sub-optimality gaps defined by $\Delta_i = 
    \mu^* - \mu_i$ and $I^*=\{i:\Delta_i=0\}$. Let 
    Assumptions \ref{ass:1} and \ref{ass:2} be fulfilled.
    There exist constants $\mathsf A_1$, $\mathsf A_2$, 
    $\mathsf A_3$ depending only on $\sigma$, $\varphi$, $\psi$, 
    $(\kappa_i,r_i, r'_i)_{i\in [m]}$, such that if the 
    parameters $\beta_{i}$ and $\rho_i$ satisfy
    \begin{align}
        \beta_{i}(n) \wedge 
        \rho_i(n) &\ge \mathsf A_1 \bar d_i
        \log\big(\mathsf A_1 m \bar d_i n/\delta\big)
    \end{align}
    for every $i\in[m]$ and every $n\in\mathbb N^*$, 
    then regret \eqref{eq:reg_def} of UCB-BO 
    \Cref{alg:coca_bo} applied up to horizon $T$, 
    with probability at least $1-2\delta$, satisfies 
    for all $T\in\mathbb N^*$ the following inequalities 
    \begin{align}
        R_T &\leq \mathsf A_2\big\{\sqrt{mT}\,\big(
        \sqrt{\beta_T\bar \gamma_T} + \rho_T\big) 
        \big\} + \|\boldsymbol{\Delta}\|_1,
        \label{eq:reg_bound_1}\\
        R_T &\le \mathsf A_3\Big\{\sqrt{|I^*| T}\,
        \big(\sqrt{\beta_T\bar \gamma_T} + \rho_T
        \big) + \big(\beta_T \gamma_T + \rho^2_T
        \big)\sum_{i\not\in I^*} \frac{1}{\Delta_i} 
        + m\big(\sqrt{\beta_1} + \rho_T\big)
        \Big\} + \|\boldsymbol{\Delta}\|_1.
        \label{eq:reg_bound_2}
    \end{align}
\end{theorem}

We now provide several comments on the upper bounds 
in \Cref{th:1}. First, the term $\|\boldsymbol{\Delta}\|_1$ 
in the regret bounds is unavoidable. It arises because 
the algorithm plays each of the $m$ arms independently 
during the initial $m$ rounds. However, since this term 
is $O(m)$ and independent of $T$, its impact on the 
asymptotic regret is negligible. Aside from this $O(m)$ 
term, the bound in \eqref{eq:reg_bound_1} is independent 
of the functions $(f_1,\ldots,f_m)$; we therefore refer 
to it as a worst-case bound. For kernels with exponential
eigendecay (e.g., the squared exponential kernel), the 
parameter $\gamma_T$ is at most polylogarithmic in $T$ 
(see \Cref{tab:sum_gamma} in the supplementary material). 
Hence, with appropriate tuning, the worst-case regret is 
of order $\sqrt{mT} \max_i \bar d_i$, up to polylogarithmic 
factors in $m$ and $T$. Notably, this yields a $\sqrt{T}\, 
\text{polylog}(T)$ dependence on $T$, a significant 
improvement over a linear regret.

One might expect the regret dependence on $m$ to improve 
beyond $\sqrt{m}$ when there is a significant gap between 
the optimal and sub-optimal arms. This is indeed the case, 
as captured by the instance-dependent bound in 
\eqref{eq:reg_bound_2}. Under mild assumptions on the 
Gaussian processes (see \Cref{sec:arm_uniqueness}), the 
set of optimal arms $I^*$ is almost surely a singleton. 
In such cases, the leading term of \eqref{eq:reg_bound_2} 
is of order $\sqrt{T} \max_i \bar d_i$ (up to polylog 
factors), improving upon the worst-case bound by a factor 
of $\sqrt{m}$. This improvement holds when the 
sub-optimality gaps $\Delta_i$ for $i \notin I^*$ are 
bounded away from zero; otherwise, the term 
$\sum_{i \notin I^*} 1/\Delta_i$ may dominate.

To the best of our knowledge, \Cref{th:1} presents the 
first regret bound not only for our proposed 
contextual-causal BO framework, but also for the more 
established Causal BO (CaBO) setting.

\raggedbottom
\section{Related Work} 
\label{sec:5}

The integration of causal knowledge to improve policy
learning is an active area of research
\citep{lee-2018, lee-2019, lee-2020}, with recent
progress on both conditional and unconditional policies
\citep{aglietti-2020, zhang-2017, zhang2022online}.
Our work builds on and extends this literature. Below,
we summarize the most related lines of work and
contrast them with our contributions.

Causal Bayesian Optimisation \citep{aglietti-2020}
uses the causal graph and a causal acquisition function
(cAF) to search over policy scopes, treating
non-intervenable variables as unobserved. In contrast,
our method explicitly leverages contextual variables and
searches the joint space of intervenable and contextual
variables. This not only leads to better solutions but
also overcomes the instabilities of cAF in contextual
scenarios. Contextual Bayesian Optimisation conditions
interventional choices on observed contexts but typically
treats all controllable variables as intervenable.
This can prevent recovery of the optimal policy. By
searching over the set of POMPSs, our method avoids
this suboptimality.

Functional Causal BO \citep{gultchin2023functional} 
places a GP prior on $\mu(\pi)$ via a finite RKHS 
expansion, $\pi(\cdot)=\sum_{i=1}^{N}\alpha_i 
k'_{\mathcal{S}}(\mathbf{c}^i,\cdot)$. Although this 
permits distance-based kernels, the critical choices 
of $N$, $\alpha_i$, and $\mathbf{c}^i$ are unspecified, 
hindering reproducibility. Furthermore, the 
optimisation of acquisition functions within this 
framework is also not addressed. In contrast, our 
method requires no such finite expansion, avoids the 
associated optimisation overhead, provides theoretical 
guarantees, and is accompanied by a public 
implementation tested on diverse environments.

Model-Based Causal BO \citep{sussex2023} assumes
specific structural causal models ($V=f_V(\mathbf{PA}_V)
+U_V$) with restrictions on the noise terms and requires
no unobserved confounders. Without any reduction of the 
search space, it considers only a limited family of 
interventions representable as parameter perturbations 
of the structural functions $f_V$. \citet{sussex2023} 
prove that $R_T \le \mathcal{O}\big(
|\mathbf{V}|\,(K\beta_T)^N\sqrt{T\gamma_T}\big)$, where 
$K$ is the maximum in-degree, $N$ is the length 
of the longest path to $Y$, $\beta_T$ and $\gamma_T$ are
similar to those defined in \Cref{main_th_sec}. 
Note that both the factor $(K\beta_T)^N$ and 
$|\mathbf{V}|$ can be very large (see \Cref{big_sys}). 
In contrast, our theory applies under more 
general conditions and the bounds scale more favorably; 
in the setting with no unobserved confounders, the set 
of POMPS is a singleton \citep{lee-2020}, leading to 
significantly improved scaling.

The method of \citep{zhang2022online} also optimises over POMPS. However, it restrictively assumes that the domain $\mathfrak{X}_{\mathbf{V}}$ is discrete and finite, an assumption not required by our approach. The regret bound established for $\text{causal-UCB}^*$ is sublinear in $T$, but it scales at least as the sum of the square roots of the cardinalities of the domains of the variables in $\mathbf V$.

{
\color{royalblue}

Another MAB-based method that utilises causal knowledge is proposed by \cite{lattimore-2016}. It is the first to show that a causal bandit framework subsumes the classical setting. Their method enjoys favourable regret bounds and is computationally efficient. As with \cite{zhang2022online}, it assumes all variables have finite support. However, it does not consider contextual interventions and assumes the absence of unobserved confounding.

Multitask GPs (\cite{bonilla2007multi}) learn similarities across the outputs of multiple functions and are useful for optimisation over multiple tasks that share inputs. \cite{aglietti-2020-multi} formalise when information sharing is possible under a known causal graph: tasks differ in their sets of interventional variables but share the same target variable. They show that sharing is possible only if every variable confounded with the target does not simultaneously have unconfounded incoming and outgoing edges. However, they study optimisation without contextual variables, and their information sharing mechanism requires computing high-dimensional integrals. Identifying assumptions that enable information sharing between POMPS without strong restrictions, and a computationally efficient mechanism to realise it, remains an open problem.

}

\begin{figure}[t]
    \centering
    \vspace*{-15pt}
    \subfloat[\tiny CaBO vs CoCa-BO (I.)]{%
        \label{fig:aspirin_statin_hom}%
        \centering
        \includegraphics[width= 0.24\textwidth ]{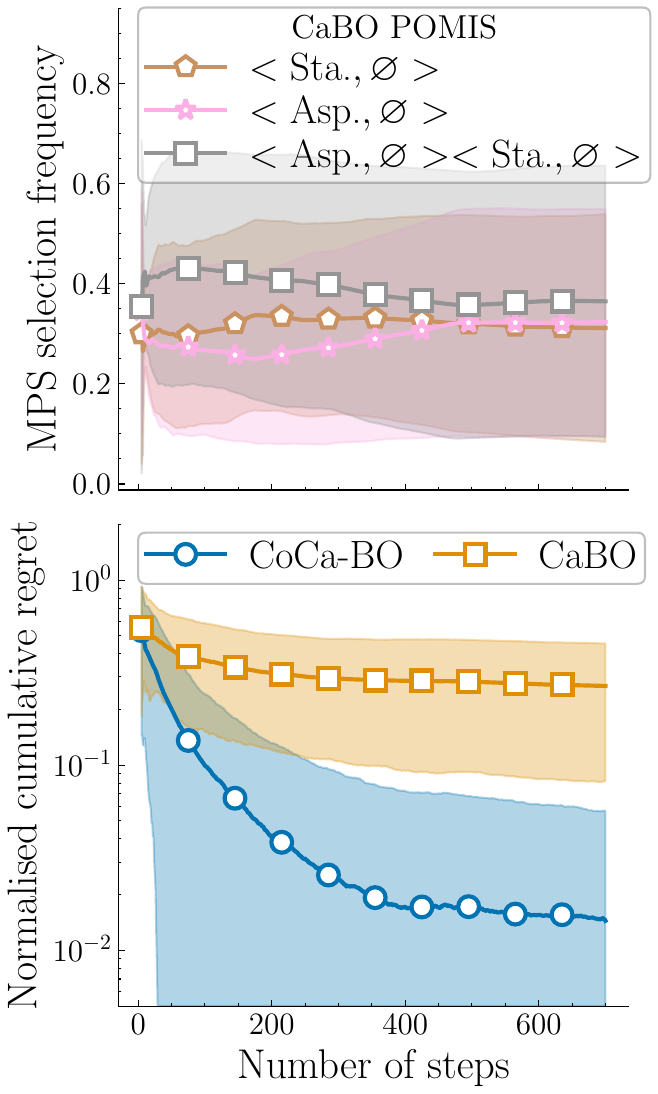}
    }
    \subfloat[\tiny CaBO vs CoCa-BO (II.)]{%
        \label{fig:aspirin_statin_het}%
        \centering
        \includegraphics[width= 0.24\textwidth ]{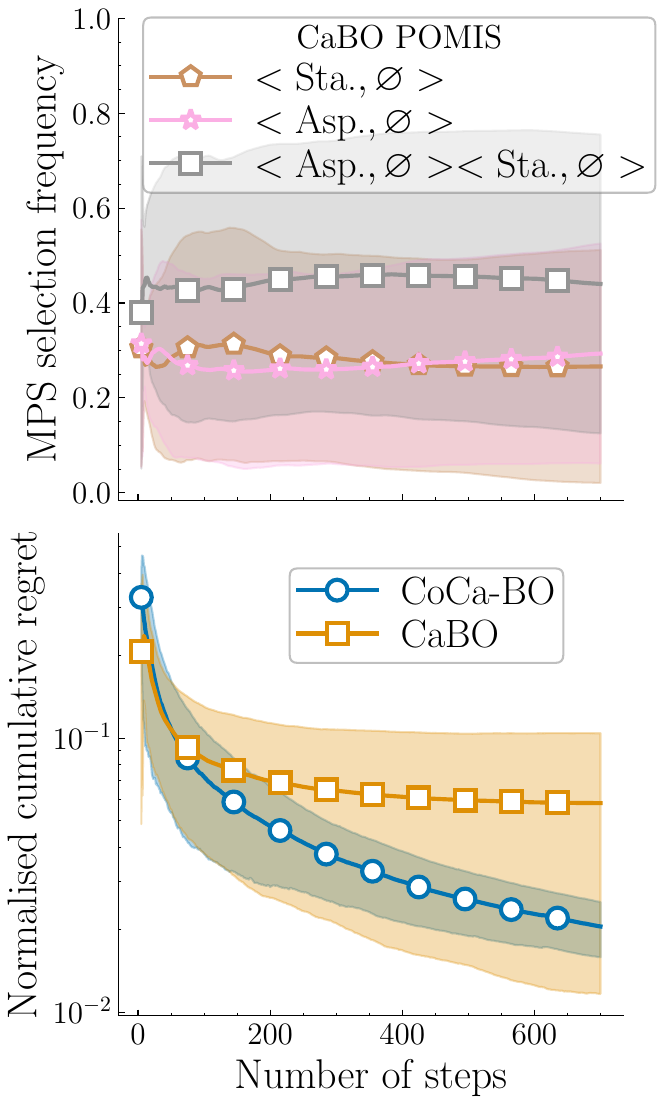}
    }
    \subfloat[\tiny  CoBO vs CoCa-BO (I.)]{%
        \label{fig:cobobetter}%
        \centering
        \includegraphics[width= 0.24\textwidth ]{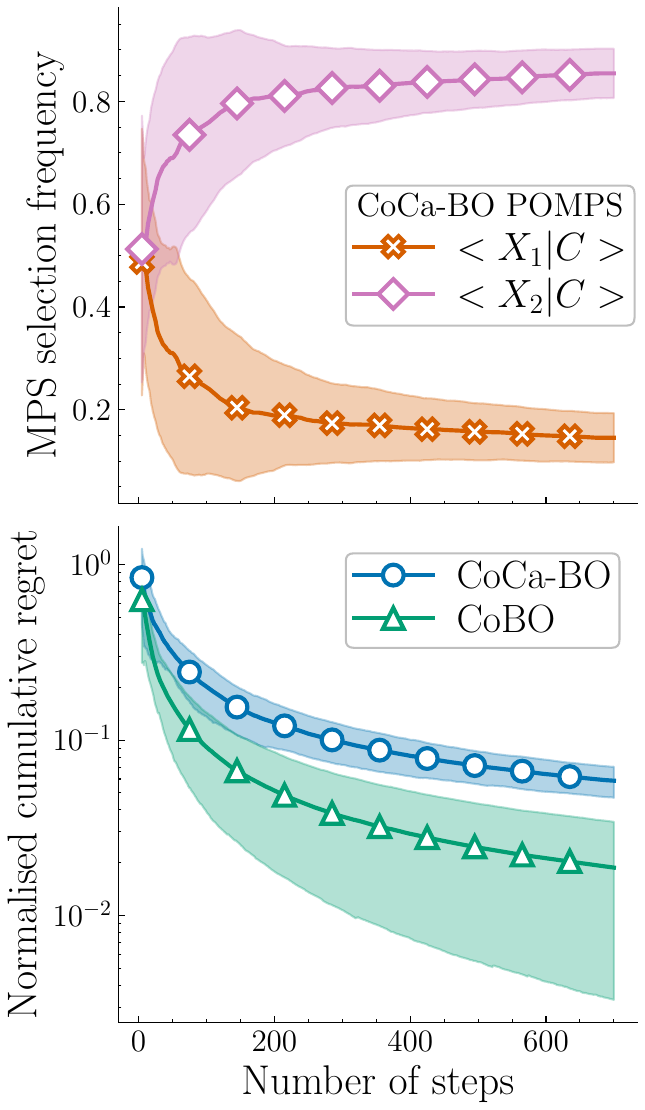}
    }
    \subfloat[\tiny  CoBO vs CoCa-BO (II.)]{%
        \label{fig:CoBOImp}%
        \centering
        \includegraphics[width= 0.24\textwidth ]{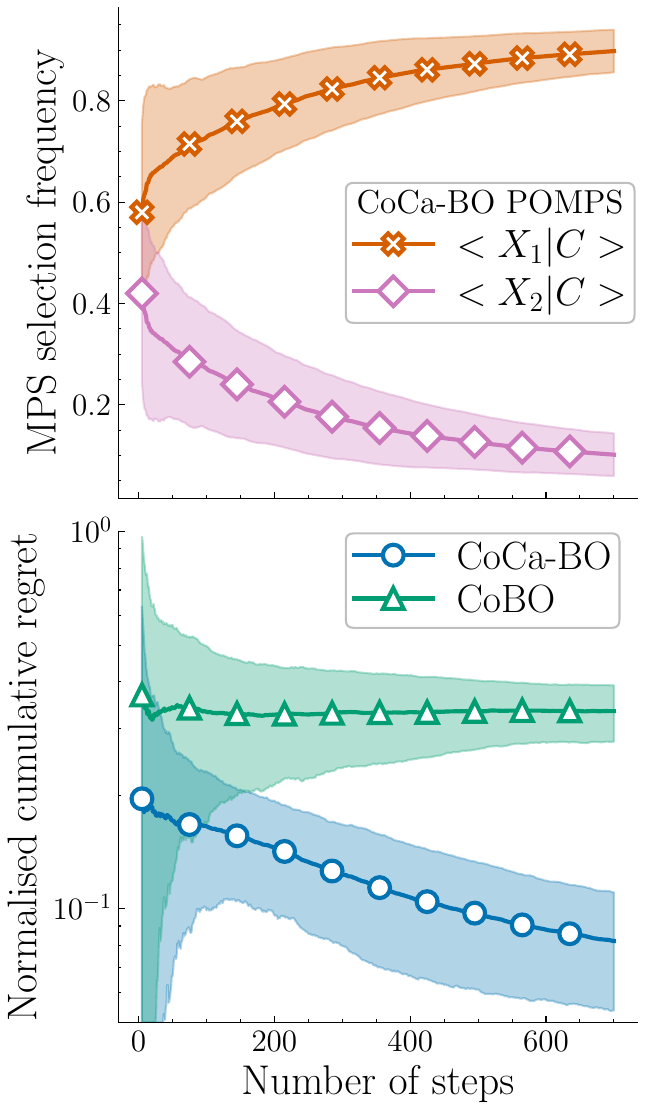}
    }     
    \caption{Top: frequency of selecting the 
    corresponding POMPS  or POMIS (omitted when there 
    is only one candidate). Bottom: 
    time-normalised cumulative regret $\bar{R}_T$.}
    \label{fig:big}
\end{figure}

\section{Experiments}\label{sec:experiments}

We evaluate the proposed method against CaBO and CoBO across a series of experiments. For each experiment and optimiser, we report the time-normalised cumulative regret
$\bar{R}_T = \frac{1}{T}\sum_{t=1}^T \hat{r}_t$
where $\hat{r}_t = \mathbb{E}_{\pi^*}[Y] - y_t$ is the immediate pseudo-regret at iteration $t$. We also track the cumulative MPS selection frequency, i.e., the fraction of times each MPS has been selected up to iteration $t$. We consider two configurations.

\vspace*{-7pt}
\begin{description}[leftmargin=1em]
    \item[\textbf{Configuration I:}] Both the proposed 
    method and the benchmark method can achieve the 
    optimum. We aim to assess the additional cost 
    introduced by the proposed method in terms of sample 
    complexity. In environments with many observed 
    variables, CoBO results in an expansive optimisation 
    domain, requiring a substantial sample size. Our method 
    leverages the independencies encoded within the 
    causal graph to reduce the optimisation domain. For 
    instance, in another example reported in 
    Appendix~\ref{big_sys}, the dimensionality of the 
    domain is reduced by a factor of 22.
    \item[\textbf{Configuration II:}] The benchmark method 
    cannot achieve the optimum. We empirically confirm that 
    there are cases where neither CoBO nor CaBO converges to
    the optimum, leading to linear cumulative regret. In the 
    same cases, we demonstrate that CoCa-BO converges to the 
    optimum policy and has sub-linear cumulative regret.
\end{description}
All experiments were conducted with 110 random seeds over 
700 iterations, corresponding to 700 distinct interventions. 
We used a single-node system with 120 CPUs, though the 
experiments can also be reproduced on a single-core 
machine with 16 GB RAM. Results are reported as averages, 
with error bars denoting 1.96 standard errors.

\vspace*{-8pt}
\paragraph{CaBO vs CoCa-BO}

\begin{figure}
\begin{floatrow}
\ffigbox{%
  \centering   
    \resizebox{.37\textwidth}{!}{
    \begin{tikzpicture}[every node/.style={circle, draw, 
    minimum size=1.5cm}]
      \node[fill=gray!30, ] (age) at (-1, 3.5) {Age};
      \node[fill=gray!30, ] (bmi) at (-1, .5) {BMI};
      \node[, ] (aspirin) at (1.5, 3) {Aspirin};
      \node[, ] (statin) at (1.5, 1) {Statin};
      \node[fill=gray!30, ] (cancer) at (3.5, 2) {Cancer};
      \node[pattern=dots] (psa) at (6, 2) {PSA};
    
      \draw[>=triangle 45, ->] (age) -- (bmi);
      \draw[>=triangle 45, ->] (age) -- (aspirin);
      \draw[>=triangle 45, ->] (bmi) -- (aspirin);
      \draw[>=triangle 45, ->] (age) -- (statin);
      \draw[>=triangle 45, ->] (bmi) -- (statin);
      \draw[>=triangle 45, ->] (aspirin) -- (cancer);
      \draw[>=triangle 45, ->] (aspirin) -- (cancer);
      \draw[>=triangle 45, ->, in=150, out=10] (aspirin) to (psa);
      \draw[>=triangle 45, ->] (cancer) to (psa);
      \draw[>=triangle 45, ->, in=-150, out=-10] (statin) to (psa);
      \draw[>=triangle 45, ->] (statin) to (cancer);
      \draw[>=triangle 45, ->, in=120, out=20] (age) to (psa);
      \draw[>=triangle 45, ->, in=-120,out=-20] (bmi) to (psa);
    \end{tikzpicture}
    }
}{%
  \caption{Causal graph of PSA level. White nodes:  
    intervenable variables; gray nodes: observable 
    variables; shaded node: target variable.}%
 \label{Fig:AspStat}
 \vspace*{-8pt}}
\ffigbox{%
  \centering
    \vspace*{-10pt}
    \includegraphics[width=.6\linewidth, height=0.45\linewidth
    ]{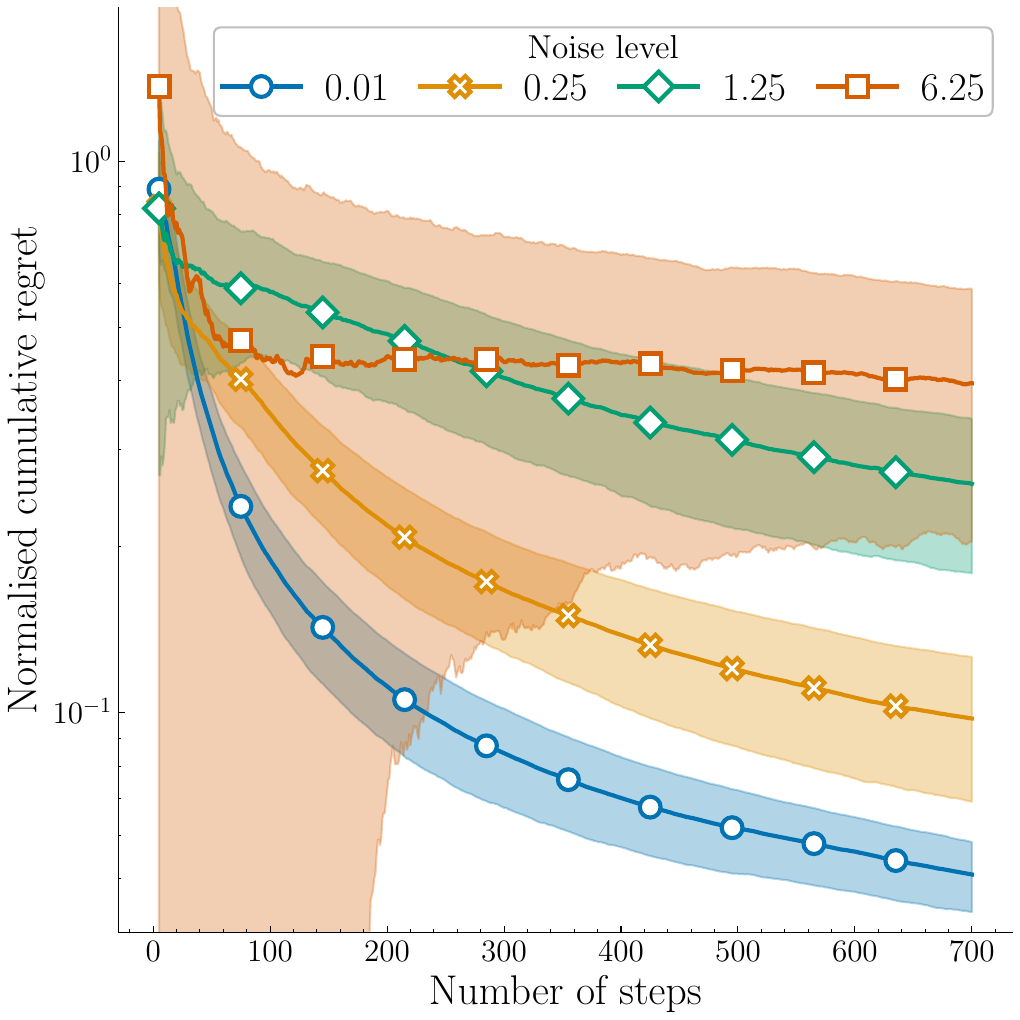}
    \vspace{-8pt}
}{%
  \caption{Robustness to the noise: Normalised 
    regret under varying noise levels.
    \label{fig:correlation}}%
}
\end{floatrow}
\vspace*{-10pt}
\end{figure}

To compare CaBO and CoCa-BO under the two aforementioned 
configurations, we consider two SCMs based on the graph of 
Fig.~\ref{Fig:AspStat} below \citep{ferro-2015,Thompson2019CausalGA}.

In this example, the variables PSA, age, BMI, cancer, 
aspirin and statin are observable, but only the last two
are intervenable. Prostate-specific antigen (PSA) 
\citep{wang-2011}, used to detect prostate cancer, is 
our target. Here, the set of POMISs is $\{\varnothing, 
\{\text{Aspirin}\}, \{\text{Statin}\}, \{\text{Aspirin, 
Statin}\}\}$ and there is only one POMPS, namely 
$\{ (\text{Aspirin}; \text{Age, BMI}); (\text{Statin}; 
\text{Age, BMI})\}$. 

\textit{First Configuration}: 
When SCM is such that the optimal value of the target 
is the same across all values of the contextual variables, 
CaBO can attain the optimum. This requires specific 
alignment of the SCM parameters, which may be unstable 
and sensitive to small fluctuations 
\citep[Ch.~2]{pearl-2009}. 
Such an SCM is defined in \Cref{cabovsus_1}. 
The optimal values at which Statin and Aspirin are 
controlled to minimise PSA are the same for all 
values of Age and BMI, that is $\forall(\textnormal{
age}, \textnormal{bmi})\in\mathfrak{X}_{\textnormal{
Age}} \times \mathfrak{X}_{\textnormal{BMI}}$:
\begin{align}
    \underset{(\text{aspirin}, \textnormal{statin})}
    {\operatorname{argmin}} \mathbb E[Y \mid & 
    \operatorname{do} (\textnormal{aspirin}, 
    \textnormal{statin}), \textnormal{age}, 
    \textnormal{bmi}] = \underset{(\textnormal{
    aspirin}, \textnormal{statin})}{\operatorname{
    argmin}}\mathbb E[Y \mid \operatorname{do}( 
    \textnormal{aspirin}, \textnormal{statin})].
\end{align}
We show that the minimum above is attained for 
aspirin$=0$  and statin$=1$.
Empirical results for this SCM appear in the first column of 
Fig.~\ref{fig:aspirin_statin_hom}. Since CaBO selects POMISs 
via the causal acquisition function, the top plot shows that 
its choices are highly variable, which harms optimiser 
convergence (bottom plot). In contrast, CoCa-BO achieves much 
smaller regret, as illustrated in the bottom plot.
    
\vspace*{-2pt}    
\textit{Second Configuration}:
The SCM for the second configuration, defined in 
Appendix~\ref{cabovsus_2}, is consistent with the causal 
graph in Fig.~\ref{Fig:AspStat}. It implies that controlling 
Aspirin and Statin based on Age and BMI yields a lower expected 
PSA than fixing them to constant values. 
Fig.~\ref{fig:aspirin_statin_het} 
depicts the results for this setup. Since CaBO marginalises 
the contextual variables, it cannot control medications based 
on Age and BMI, and therefore fails to converge (bottom plot). 
The top plot shows again the high variability in POMIS 
selection under CaBO. As expected, CoCa-BO attains a much 
smaller regret.

\vspace*{-8pt}
\paragraph{CoBO vs CoCa-BO}
We consider two SCMs based on the graph of 
Fig.~\ref{Fig:1}. 

\textit{First Configuration:}
    Appendix~\ref{cobo_good} describes an SCM consistent 
    with the causal graph in Fig.~\ref{Fig:1}, where 
    intervening on both $X_1$ and $X_2$ does not hinder the 
    agent's ability to reach the optimal outcome. 
    Fig.~\ref{fig:cobobetter} (bottom plot) shows both 
    methods converge to the optimum, as evidenced by the 
    decreasing normalised cumulative regret, with CoBO 
    converging faster than CoCa-BO. 
    In this configuration, CoCa-BO's advantage 
    over CoBO emerges in environments with many variables,
    as shown in \Cref{big_sys}.
    
    
\textit{Second Configuration:}
    The second configuration examined in this study is the 
    SCM of Fig.~\ref{Fig:1}. We have shown analytically that 
    the CoBO method is unable to attain the optimal solution. 
    Our method, CoCa-BO, has been demonstrated to 
    effectively achieve the desired optimum, resulting in 
    sub-linear regret. The top sub-figure of 
    Fig.~\ref{fig:CoBOImp} illustrates the cumulative
    frequency of selecting the corresponding POMPS. We 
    observe that CoCa-BO converges to the policy scope 
    $(X_1; C)$ yielding the optimal value. The lower 
    sub-figure of Fig.~\ref{fig:CoBOImp} 
    illustrates that the normalised cumulative regret remains 
    constant for CoBO, indicating its linear cumulative regret. 
    On the other hand, CoCa-BO attains sub-linear cumulative 
    regret, as evidenced by the decreasing trend of the blue
    curve.

\vspace*{-8pt}
\paragraph{Robustness to Noise}

The target variable $Y$ is pivotal, as its observed 
values guide both policy scope and selection. 
Consequently, higher noise in $Y$ slows the convergence of 
optimisation algorithms \citep{Lattimore_Szepesvári_2020,srin_kauss_cont,srinvas_kauss}. 
To assess the robustness of CoCa-BO to noise in $Y$, we 
vary the standard deviation of $\epsilon_Y$ in the 
example of Appendix~\ref{cobo_good}, adjusting 
$\sigma(\epsilon_Y)$ from $0.01$ to $6.25$. As shown in 
Fig.~\ref{fig:correlation}, CoCa-BO converges for 
$\sigma$ up to $1.25$, while convergence becomes unclear 
under extreme noise, e.g., $\sigma(\epsilon_Y)=6.25$.

\section{Conclusion}

We provided evidence that existing Bayesian optimisation methods that incorporate causal knowledge are not capable of integrating contextual information without incurring substantial computational and statistical costs. To overcome these challenges, 
we introduced CoCa-BO, a novel method that efficiently optimises over mixed policy scopes. We derived worst-case and instance-dependent regret bounds, which also address theoretical gaps in previous methods. We conducted experiments across various environments and demonstrated that CoCa-BO consistently reached the optimum even where existing methods could not. Moreover, our experiments showed that CoCa-BO is robust to noise and reduces sample complexity by identifying smaller optimisation domains in environments with a large number of variables. To support reproducibility, we provide our Pyro-based implementation \citep{bingham2018pyro, phan2019composable}, offering a unified framework for future research.

\section*{Acknowledgements} 
This work was supported by Hi! PARIS and received government 
funding managed by the Agence Nationale de la Recherche under 
the France 2030 program, references (ANR-23-IACL-0005) and
(ANR-23-PEIA-0004), 
as well as 
by ERC grant SAGMOS (grant agreement No. 101201229).

\bibliography{example_paper}
\bibliographystyle{icml2025}

\newpage
\onecolumn
\appendix
\renewcommand\contentsname{}
\part{Appendix} 
\parttoc 

\newpage

\begin{table}[ht]
    \centering
    \begin{tabular}{l|l}
        \toprule
        Acronym & Full text\\
        \midrule
        BO & Bayesian Optimisation \\
        CaBO & Causal Bayesian Optimisation\\
        CoBO & Contextual Bayesian Optimisation\\
        CoCa-BO & Contextual Causal Bayesian Optimisation\\
        POMIS & Possibly Optimal Minimal Intervention Set\\
        SCM & Structural Causal Model\\
        GP & Gaussian Process\\
        UCB & Upper Confidence Bound\\
        DAG & Directed Acyclic Graph\\
        MPS & Mixed Policy Scope\\
        POMPS & Possibly-Optimal Mixed Policy Scope\\
        MAB & Multi-Armed Bandit\\
        \bottomrule
    \end{tabular}
    \caption{This table presents the main acronyms used throughout the paper.}
    \label{tab:1}
\end{table}


\begin{table}[ht]
    \centering
    \begin{tabular}{l|p{10cm}}
        \toprule
        Notation & Meaning\\
        \midrule
        $\mathbf U,\mathbf V,\mathbf X$ & Vectors/sets composed of 
        random variables\\[3pt]
        $\mathbf u,\mathbf v,\mathbf x$ & Values taken by vectors/sets 
        of random variables\\[3pt]
        $U,V,X$ & random variables\\[3pt]
        $\mathfrak{X}_U,\mathfrak{X}_V,\mathfrak{X}_X$ & The domains 
        of the random variables $U,V,X$; \textit{i.e.}, the sets in which 
        $U,V,X$ take their values. \\[3pt]
        $\mathfrak{X}_{\mathbf U},\mathfrak{X}_{\mathbf V}, 
        \mathfrak{X}_{\mathbf X}$ & The domains 
        of the random vectors $\mathbf U, \mathbf V,\mathbf X$. \\[3pt]
        $P_{\mathbf V}^{\square}$ & A probability distribution of 
        the random vector $\mathbf V$; $\square$ can be replaced by 
        different symbols, for instance, by $0$ or by $\pi$.\\[3pt]
        $\mathbb E_{P^\square_{\mathbf V}}[Z]$ & 
        The expectation of the random variable $Z = g(\mathbf V)$ 
        assuming that $\mathbf V$ is drawn from $P_{\mathbf V}^\square$. \\[3pt]
        $\mathcal F$ & Set of functions with elements usually denoted by $f$.\\[3pt]
        $SCM$ & A structural causal model given by the quadruple $(\mathbf 
        V,\mathbf U,\mathcal F,P_{\mathbf U})$.\\[3pt]
        $\mathcal G$ & A directed acyclic graph.\\[3pt]
        $\mathbf{PA}_V$ & When $V$ is a node in a DAG, $\mathbf{PA}_V$
        is the set of its parent nodes.\\[3pt]
        $\mathbf{C}_X$ & When $X$ is an intervenable variable, 
        $\mathbf{C}_X$ is its scope, \textit{i.e.}, a set of observable 
        variables that can be used for planning an intervention on $X$. \\[3pt]
        $\mathcal S$ & A mixed policy scope, \textit{i.e.}, a collection 
        of pairs $(X,\mathbf{C}_X)$. \\[3pt]
        $\mathbf X(\mathcal S)$ &  The set of intervenable variables of 
        $\mathcal S$.\\[3pt]
        $\mathbf C(\mathcal S)$ &  The set of contextual variables of 
        $\mathcal S$.\\[3pt]
        $\mathcal{G}_{\mathcal{S}}$ & The graph obtained from $\mathcal G$ 
        by removing edges into $X$  and adding new ones from each element 
        of $\mathbf{C}_X$ to $X$ for every $(X,\mathbf{C}_X)\in\mathcal S$. \\[3pt]
        $\pi_{X|\mathbf{C}_X}$ & a mapping from $\mathfrak{X}_{\mathbf{C}_X}$ 
        to $\mathfrak{X}_X$.\\[3pt]
        $\mathbb S[\mathcal G]$ & All mixed policy scopes $\mathcal S$ compatible 
        with the DAG $\mathcal G$.\\[3pt]
        $\pi$ & A mixed policy, \textit{i.e.}, a collection of 
        mappings $\pi_{X|\mathbf{C}_X}$\\[3pt]
        $\Pi_{\mathcal S}$& The set of all mixed policies $\pi$ based on 
        the mixed policy scope $\mathcal S$. \\[3pt]
        $\Pi_{\mathcal G}$& The set of all mixed policies $\pi$ based on 
        a mixed policy scope $\mathcal S\in \mathbb S[\mathcal G]$. \\[3pt]
        $\SCM^\pi$ & The SCM obtained from $\SCM$ by applying the 
        policy $\pi$. \\[3pt]
        $\mu(\pi)$ & The expected reward under the mixed policy $\pi$. 
        For a well specified $Y\in\mathbf V$, it is defined as $\mu(\pi) = 
        \mathbb E_{P_{\mathbf V}^\pi}[Y]$. \\[3pt]
        $\mu^*_{\mathcal S}$ & The highest expected reward among all 
        mixed policies $\pi$ based on the scope $\mathcal S$.\\[3pt]
        $\mathbb E_{\pi}[Z]$ & For a random variable $Z = g(\mathbf V)$ 
        and a mixed policy $\pi$, this is a shorthand for $\mathbb 
        E_{P_{\mathbf V}^\pi}[Z]$.\\[3pt]
        $\mathbb S^*[\mathcal G]$ & The subset of $\mathbb S[\mathcal G]$
        consisting of only those $\mathcal S$ that are possibly optimal.\\[3pt]
        $\mathcal D^t$ & Historical data up to time $t\in\mathbb N$, 
        consisting of quadruplets $(\mathcal S_i,\mathbf c_i,\mathbf x_i,y_i)$
        for $i=1,\ldots,t$.\\[3pt]$\mu^{(i)}_t, 
        \sigma^{(i)}_t$ & Posterior mean and standard 
        deviation of the $i^{th}$ Gaussian process after 
        time $t$.\\[3pt]
        $\hat{r}_t,R_t, \bar{R}_t$ & Instantaneous regret 
        at time $t$, cumulative regret at time $t$, and 
        time-normalised cumulative regret at time $t$.\\[3pt]
        \bottomrule
    \end{tabular}
    \caption{This table summarises the main notations used throughout the paper.}
    \label{tab:2}
\end{table}

\section{Additional details on the experimental results}

In this section, we present the details necessary to 
understand and reproduce the experiments reported 
in \Cref{sec:experiments}.

\subsection{CaBO vs CoCa-BO}
We describe and analyse SCMs used for comparing CaBO with CoCa-BO in this section.

\subsubsection{First Configuration}
\label{cabovsus_1}
The SCM that is used for the comparison of CaBO vs CoCa-BO in a scenario where CaBO can achieve the optimum is given below:
$\text {Age}\sim\mathcal{U}(55,75)$, $\epsilon  \sim 
\mathcal{N}(0, 0.4)$ and\footnote{Here, $\varrho$ denotes the sigmoid function.}
\begin{align}
\text { BMI } & \sim\mathcal{N}(27.0-0.01 \times\text { Age }, 0.7) \\
\text { Aspirin } & =\varrho(-8.0+0.10 \times \text { Age }+0.03 \times \text{BMI}) \\
\text { Statin } & =\varrho(-13.0+0.10 \times \text { Age }+0.20 \times \text{BMI}) \\
\text { Cancer } & =\varrho(2.2-0.05 \times \text { Age }+0.01 \times \text{BMI}-  0.04 \times  \text { Statin }+0.02 \times \text { Aspirin }) \\
\text{PSA} & = \epsilon+6.8+0.04 \times \text { Age }-0.15 \times \text{BMI}- 0.60 \times \text { Statin }+0.55 \times \text { Aspirin }+\text { Cancer }
\end{align}
We first note that for any values of Age and BMI 
\begin{align}
    \mathbb{E}[\text{PSA}&\mid \operatorname{do}(\textnormal{asp}, \textnormal{sta}), \textnormal{age}, \textnormal{bmi}]\\
    &=\mathbb{E}[\epsilon+6.8+0.04 \text { Age }-0.15  \text{BMI}-0.60  \text { Sta }+0.55  \text { Asp }+\text { Can }\mid \operatorname{do}(\textnormal{asp}, \textnormal{sta}), \textnormal{age}, \textnormal{bmi}]\\
    &=
    6.8+0.04\textnormal{age}-0.15 \textnormal{bmi}-0.6 \textnormal{sta} + 0.55 \textnormal{asp}+\mathbb{E}[\text { Can }\mid \operatorname{do}(\textnormal{asp}, \textnormal{sta}), \textnormal{age}, \textnormal{bmi}]\\
    & = 6.8 + 0.04 \textnormal{age} - 0.15 \textnormal{bmi} - 0.6 \textnormal{sta} + 0.55 \textnormal{asp}\\ 
    &\qquad + \varrho(2.2 - 0.05 \textnormal{age} + 0.01 \textnormal{bmi} - 0.04 \textnormal{sta} + 0.02 \textnormal{asp})
\end{align}
It is easy to see that
\begin{align}
    \underset{\textnormal{asp}, \textnormal{sta}}{\operatorname{argmin}}\,&\mathbb{E}[\text{PSA}\mid \operatorname{do}(\textnormal{aspirin}, \textnormal{statin}), \textnormal{age}, \textnormal{bmi}]  =\\
    &\underset{\textnormal{asp}, \textnormal{sta}}{\operatorname{argmin}} -0.6 \  \textnormal{sta} + 0.55\ \textnormal{asp}+\varrho(2.2 - 0.05 \textnormal{age} + 0.01\textnormal{bmi} - 0.04 \textnormal{sta} + 0.02\textnormal{asp}).
\end{align}
In summary, to minimise PSA, Statin should be controlled at its highest value and Aspirin at its lowest value. Since the domains of both variables are $[0, 1]$, the solution of the optimisation problem is $\text{Aspirin}=0$ and $\text{Statin}=1$, which is independent of Age and BMI. This implies that the optimal control values for Aspirin and Statin are the same for all values of Age and BMI. This is why the optimum is achievable by CaBO.

\subsubsection{Second Configuration}
\label{cabovsus_2}
The SCM that is used for this setup is such that an agent should control Aspirin and Statin based on Age and BMI to achieve the minimum expected value for PSA. It is defined by
$\text { Age } \sim\mathcal{U}(55,75) $, $\epsilon 
\sim \mathcal{N}(0, 0.01)$ and 
\begin{align}
\text { BMI } & \sim\mathcal{N}(27.0-0.01 \times\text { Age }, 0.1) \\
\text { Aspirin } & =\varrho(-8.0+0.10 \times \text { Age }+0.03 \times \text{BMI}) \\
\text { Statin } & =\varrho(-13.0+0.10 \times \text { Age }+0.20 \times \text{BMI}) \\
\text { Cancer } & = \text{Statin}^2 + \text{AgeBMI}^2 
    + 0.5\times \text{Aspirin}^2 \\
\text{PSA} & = \epsilon+ \text{Cancer} + 0.5\times \text{Asp}^2 +  
    \text{AgeBMI}^2 - 2\times\text{AgeBMI}\times(\text{Asp}
    + \text{Sta}),
\end{align}
where we used the notation
\begin{align}
    \text{AgeBMI} = \Big(\frac{\text{Age}-55}{21}\Big)
    \Big| \frac{\text{BMI}-27}{4}\Big|.
\end{align}
Plugging in the structural equation of Cancer into PSA, we get
\begin{align}
    \text{PSA}&=\epsilon + \text{Asp}^2+\text{Sta}^2 + 2
    \text{AgeBMI}^2 - 2\times\text{Asp}
    \text{AgeBMI} - 2\times \text{Sta}\times \text{AgeBMI}\\
    &=
    \left(\text{Asp}-\text{AgeBMI}\right)^2+\left(\text{Sta} 
    -\text{AgeBMI}\right)^2+\epsilon
\end{align}
An agent controlling both Aspirin and Statin at $\left(\frac{\text{Age}-55}{21}\right)\left| \frac{\text{BMI}-27}{4}\right|$ level achieves the minimal expected value of PSA, namely 0. CaBO fails as it does not consider Age and BMI while controlling Aspirin and Statin.

\subsection{CoBO vs CoCa-BO}
\label{cobo_good}
We  introduce an SCM such that CoBO which by default controls 
both $X_1$ and $X_2$ based on $C$ is still capable of getting 
to the optimum value. The SCM is based on the causal graph 
depicted in Fig.~\ref{Fig:1} and the following equations 
\begin{align}
    U_1, U_2 &\sim \operatorname{Uniform}(-1, 1)\\
    \epsilon_C, \epsilon_{X_1},  &\sim \mathcal{N}(0, 0.1), \ \epsilon_{Y}\sim\mathcal{N}(0, 0.01) \\
    C &= U_1+\epsilon_C\\
    X_1 &= U_1+\epsilon_{X_1}\\
    X_2 &= \left| C-X_1\right|+0.2U_2\\
    Y &= \operatorname{cos}(C-X_2)+0.1U_2+\epsilon_{Y}. 
\end{align}
This SCM has two POMPSs, namely $( X_1; C)$ and $(X_2; C)$.

We first note that intervening on both $X_1$ and $X_2$ is redundant in the sense that intervening on $X_2$ destroys the directed causal path from $X_1$ to $Y$ and renders the intervention on $X_1$ irrelevant to $Y$. Let the intervention on $X_2$ be given by an arbitrary function $\pi:\mathfrak{X}_C\mapsto \mathfrak{X}_{X_2}$. Then the expectation of $Y$ under such control can be expressed as:
\begin{align}
\mathbb{E}_\pi[Y]=\mathbb{E}_\pi[\operatorname{cos}(C-X_2)+0.1U_2+\epsilon_y]=\mathbb{E}_\pi[\operatorname{cos}(C-X_2)]=\mathbb{E}_\pi[\operatorname{cos}(C-\pi(C))].  
\end{align}
It is trivial to see that to maximise the above expectation, one should set $X_2$ at the value of $C$, \textit{i.e.}, $\pi$ is the identity. This yields 
$\max_\pi\mathbb{E}_\pi[
\operatorname{cos}(C-\pi(C))]=1$.

Therefore intervening on $X_2$ based on $C$ yields the optimal value, as 1 which is the highest value that the expectation of $Y$ may achieve under any policy $\mathbb{E}_\pi[Y]=\mathbb{E}_\pi[\operatorname{cos}(C-X_2)+0.1U_2+0.1\epsilon_y]=\mathbb{E}_\pi[\operatorname{cos}(C-X_2)]\leq 1$. We conclude that CoBO can achieve the optimal value for this SCM.

\subsection{Environment with a large number of variables}
\label{big_sys}

This example highlights a key advantage of CoCa-BO: 
it can yield significant gains in sample efficiency 
over standard CoBO, even when both methods converge 
to the same optimum. The advantage stems from using 
the causal graph to identify and remove superfluous 
variables from the policy scope. The resulting 
reduction in dimensionality often accelerates 
learning. We demonstrate this in a challenging, 
high-dimensional setting with 57 contextual and 
29 intervenable variables, where the improvement 
from a simplified scope outweighs the cost of 
searching over the policy space. The underlying 
Structural Causal Model is defined as follows.

\begin{figure}[h]
    \centering
    \subfloat[]{%
        \label{fig:my_label}%
        \begin{minipage}{0.5\linewidth}
            \centering
    \resizebox{\textwidth}{!} {
    \begin{tikzpicture}[]
     \node[fill=gray!30, circle, draw, minimum size=1.cm] (C0) at (0, 9) {$C_0$};
     \node[fill=gray!30, circle, draw, minimum size=1.cm] (C1) at (2, 9) {$C_1$};
     \node[fill=gray!30, circle, draw, minimum size=1.cm] (C2) at (2, 7) {$C_2$};
     \node[fill=gray!30, circle, draw, minimum size=1.cm] (C3) at (2, 5) {$C_3$};
     \node[fill=gray!30, circle, draw, minimum size=1.cm] (C4) at (3.5, 3.5) {$C_4$};
     \node[fill=gray!30, circle, draw, minimum size=1.cm] (C5) at (5, 2) {$C_5$};
     \node[fill=gray!30, circle, draw, minimum size=1.cm] (C6) at (6, 4) {$C_6$};
     \node[fill=gray!30, circle, draw, minimum size=1.cm] (Cti) at (0, 4.5) {$C_{t_i}$};
     \plate {Cti_plate} {(Cti)} {\textbf{50}}

     \node[, circle, draw, minimum size=1.cm] (X1) at (5, 8) {$X_1$};
     \node[, circle, draw, minimum size=1.cm] (X2) at (6, 6) {$X_2$};
     \node[, circle, draw, minimum size=1.cm] (Xtj) at (0, 2) {$X_{t_j}$};
     \plate {Xtj_plate} {(Xtj)} {\textbf{27}}
     
      \node[pattern=dots, circle, draw, minimum size=1.cm] (Y) at (8.5, 6) {$Y$};

  \draw[>=triangle 45, ->] (C0) to (C1);
  \draw[>=triangle 45, ->] (C1) to (C2);
  \draw[>=triangle 45, ->] (C1) to (X2);
  \draw[>=triangle 45, <->, dashed, in=130, out=10] (C1) to (X1);
  \draw[>=triangle 45, ->] (C2) to (C3);
  \draw[>=triangle 45, ->] (C3) to (C4);
  \draw[>=triangle 45, ->] (C4) to (C5);
  \draw[>=triangle 45, ->] (C5) to (C6);
  \draw[>=triangle 45, ->] (C6) to (X2);
  \draw[>=triangle 45, ->] (X1) to (X2);
  \draw[>=triangle 45, ->] (X2) to (Y);
     \draw[>=triangle 45, <->, dashed, in=35,out=145] (Y) to (X2);
  \draw[>=triangle 45, ->] (Cti) to (Xtj);
  \draw[>=triangle 45, ->] (Xtj) to (C5);
      \draw[>=triangle 45, ->, out=30, in=100] (C1) to (Y);

    \end{tikzpicture}}
    \end{minipage}
    }\hfil
    \subfloat[Performance of CoBO and CoCa-BO]{%
        \label{fig:big_example}%
    \begin{minipage}{0.3\linewidth}
        \includegraphics[width=\textwidth]{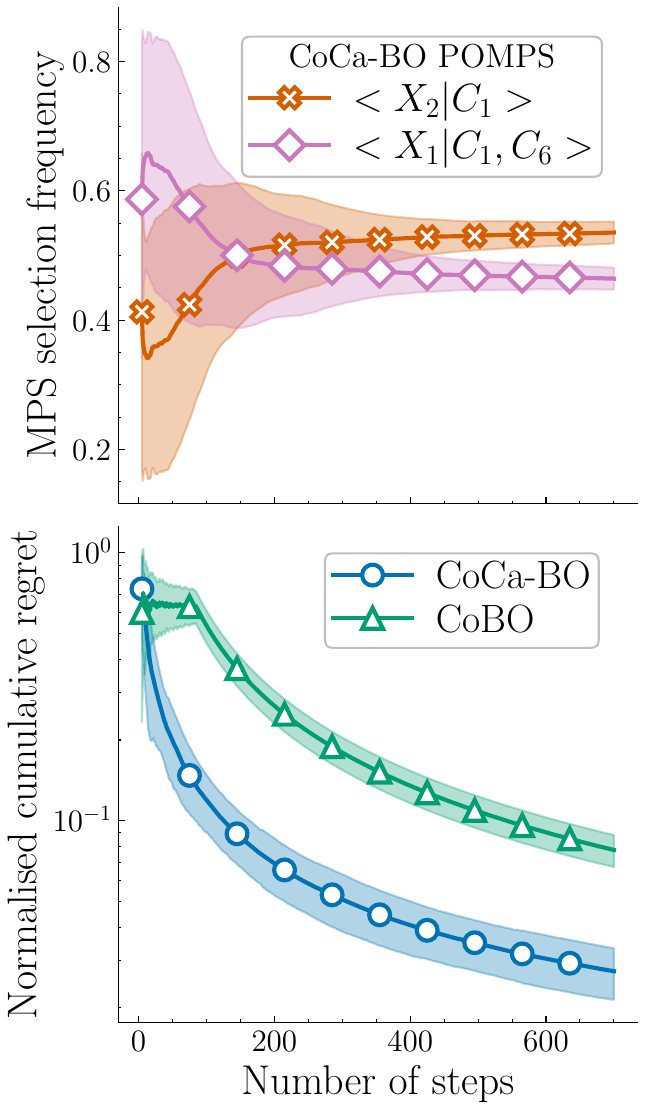}
    \end{minipage}}\hfil
    \caption{(a) A causal graph with a large number of variables. The rectangular nodes represent plates of variables. The number below the rectangles represents the number of variables within each plate. Variables $C_{t_i}$ and $X_{t_j}$ over-specify the environment and are redundant for optimising the expected value of $Y$. (b) Both methods converge, but CoCa-BO has lower  cumulative cost due to smaller policy scopes ($\{( X_1; C_1, C_6), ( X_2; C_1)\}.$
    }
    \label{fig:CoBO_hard}
\end{figure}

\begin{align}
    U_1, U_2 &\sim \operatorname{Uniform}(-1, 1),
    &C_0  &\sim \mathcal{N}(0, 0.2)& & &\\
    X_1 & \sim \mathcal{N}(U_1  , 0.1),
    &C_1  &\sim \mathcal{N}(C_0-U_1  , 0.1)& & & \\
    C_2 & \sim \mathcal{N}(C_1  , 0.1), 
    &C_3 & \sim \mathcal{N}(C_2  , 0.1)&
    C_4 & \sim \mathcal{N}(C_3  , 0.1)\\
    C_{t_i} & \sim \mathcal{N}(0, 0.2)  &\forall{i}&\in\{ 1,\dots, 50\}& \bar C_{50} &=  \frac{1}{50}\sum^{50}_{i=1}C_{t_i}\\
    X_{t_j} &\sim \operatorname{Uniform}\big[-1, 1+0.1\bar C_{50}\big]   &\forall{j}&\in\{ 1,\dots, 27\}
    & \bar X_{27} &= \frac{1}{27}\sum^{27}_{j=1}X_{t_j}\\
    C_5 & \sim \mathcal{N}\big(C_4+0.01\bar X_{27}  , 0.1\big)
    &C_6 & \sim \mathcal{N}(C_5  , 0.1) &&\\
    X_2 & \sim \mathcal{N}(0.5(C_1+C_6)+X_1+0.3\left|U_2\right|  , 0.1)&&&&\\
    Y &\sim \mathcal{N}(\operatorname{cos}(C_1-X_2)+0.1U_2, 0.01)
    &&&&
\end{align}

We also provide the causal graph corresponding to the above SCM in \ref{fig:my_label}, where we use rectangular nodes to represent the plate notation \citep{koller2009probabilistic}. This example has purely contextual variables $C_{t_i}$ and $X_{t_j}$ that only influence the rest of the system only through $C_5$. It is not necessary to control or observe these variables to achieve highest expectation of the target variable $Y$. This is reflected in the POMPSs $\{( X_1; C_1, C_6), ( X_2; C_1)\}$ corresponding the causal graph. Both POMPSs contain noticeably 
fewer variables than  the policy scope considered by CoBO $( X_1, X_2, X_{t_{1:27}};C_{0:6}, C_{t_{1:50}})$ where $X_{m:n}=\{X_i\mid  m \leq i \leq n\}$. 

One may note that this SCM is similar to the SCM  from \ref{cobo_good}. Indeed, the structural equations for $Y$ are the same. The same derivations apply to this SCM, proving that CoBO can achieve the optimum, which is reflected in Fig. \ref{fig:big_example}, which shows that both methods converge to the optimum, yielding a decreasing trend in the normalised cumulative regret. Moreover, CoCa-BO converges faster as it utilises the causal graph and derives policies with a smaller scope. As we see, this significantly improves the convergence speed over CoBO even if it achieves the optimum.

\subsection{Robustness to Noise}
\label{app:rob_un}

We note that the total variance of $Y$ under the optimal policy is actually higher than the variance of $\epsilon_Y$ due to an unobserved confounder $U_2$, which has a variance of $1/3$. Thus, $\text{var}_{\pi}(Y) = \frac{10^{-2}}{3} + \text{var}(\epsilon_Y)$
under the optimal policy $\pi$. We emphasise that $\text{var}_{\pi'}(Y)$ under any other policy $\pi'$ is greater than or equal to $\text{var}_{\pi}(Y)$.

{
\color{royalblue}
\subsection{Robustness to GP backend and Importance of Context}

This subsection examines the effect of the underlying BO optimisation on our method. We ablate the backend across HEBO (\cite{hebo}), BoTorch (\cite{botorch}), and GPflow (\cite{gpflow}). HEBO and BoTorch utilise GPyTorch (\cite{gpytorch}) for GP inference, whereas GPflow provides its own TensorFlow-based implementation (\cite{tensorflow}).

We conduct the study on the SCM described in Appendix~\ref{cobo_good} and summarise the results in \Cref{fig:backend_comp}. We observe that our method converges regardless of the underlying BO routine, though HEBO appears slightly better.

In the same environment, we ablate context by removing the contextual variables from all POMPS while preserving the interventional variables and the hierarchical design of CoCa-BO. The results in \Cref{fig:context_vs_no_context} show that, without contextual information, the algorithm quickly converges to a suboptimal solution and cannot escape it, due to the missing context.

\begin{figure}[t]
    \centering
    \vspace*{-15pt}
    \subfloat[\small BO backend comparison]{%
        \label{fig:backend_comp}%
        \centering
        \includegraphics[width= 0.4\textwidth ]{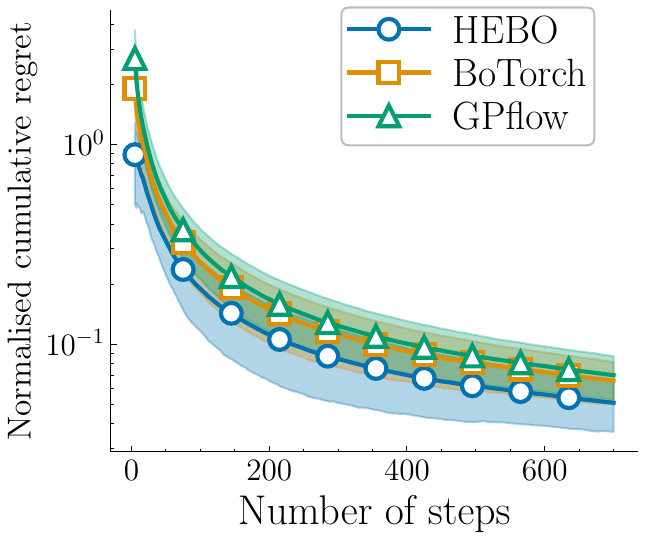}
    }
    \subfloat[\small Importance of context]{%
        \label{fig:context_vs_no_context}%
        \centering
        \includegraphics[width= 0.4\textwidth ]{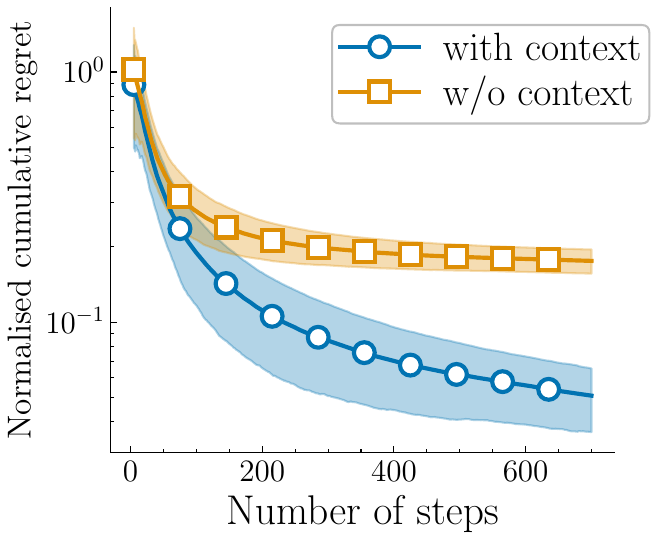}
    }    
    \caption{Left: CoCa-BO performance under different BO implementations. Right: CoCa-BO performance with contextual information preserved versus discarded.}
    \label{fig:big}
\end{figure}

}

\section{Main theoretical results}
\label{app:main}
\subsection{Description of the setting and the algorithm}
\label{ssec:setting}

Our algorithm, described in \Cref{alg:coca_bo}, does not 
distinguish between two POMPSs that share the same sets of intervenable and
contextual variables. Hence, we say that two MPSs $\mathcal 
S$ and $\mathcal S'$ are equivalent if $\mathbf{X}(\mathcal 
S) = \mathbf{X}(\mathcal S')$ and $\mathbf{C}(\mathcal S) =
\mathbf{C}(\mathcal S')$. Let $m$ denote the number of 
distinct equivalence classes of POMPSs. We index these 
equivalence classes by $i \in \{1,\dots,m\}$. For the $i$-th 
equivalence class, we denote its intervenable variables by
$\mathbf X(i)$ and its contextual variables by $\mathbf C(i)$.
We denote by $P_i$ the distribution of the random vector
$\mathbf C(i)$. 

We assume that the reward obtained under a POMPS $\mathcal 
S$ belonging to the $i$-th class is given by
\[
    Y \;=\; f_i\bigl(\mathbf{X}(i), \mathbf{C}(i)\bigr) \;+\; 
    \varepsilon_i,
\]
where $f_i : \mathfrak{X}_{\mathbf{X}(i)} \times \mathfrak{X
}_{\mathbf{C}(i)} \to \mathbb{R}$ and $\varepsilon_i$ is
independent Gaussian noise, $\varepsilon_i \sim \mathcal{N}(0,
\sigma^2)$. Furthermore, for every equivalence class, we 
assume $f_i \sim \mathcal{GP}(0, k_i)$ independently across 
$i$, where $k_i$ are known covariance functions such that $\kappa_i:=\sup_{\mathbf{x},\mathbf{c}, \mathbf{x}', \mathbf{c}'} k_i((\mathbf{x},\mathbf{c}), (\mathbf{x}',\mathbf{c}'))< \infty$.



This means that we assume that the following arrays of 
independent random objects are generated by the nature
\begin{align}
    \begin{matrix}
        i=1 :& f_1;  & (\mathbf C_{1,1}, 
        \mathbf C_{1,2}, \mathbf C_{1,3},\ldots, \mathbf 
        C_{1,\ell},\ldots) &\ \sim \mathcal{GP}(0, k_1)
        \otimes P_1^{\otimes \mathbb N^*},\\[2pt]
        i=2 :& f_2; & (\mathbf C_{2,1}, 
        \mathbf C_{2,2}, \mathbf C_{2,3},\ldots, \mathbf 
        C_{2,\ell},\ldots)& \ \sim \mathcal{GP}(0, k_2)
        \otimes P_2^{\otimes \mathbb N^*},\\[2pt]
        \vdots &  \vdots & \vdots& \vdots\\
        i=m :& f_m; & (\mathbf C_{m,1}, 
        \mathbf C_{m,2}, \mathbf C_{m,3},\ldots, \mathbf 
        C_{m,\ell},\ldots)& \ \sim \mathcal{GP}(0, k_m)
        \otimes P_m^{\otimes \mathbb N^*},\\[4pt]
        & & (\varepsilon_1,\varepsilon_2,\ldots,\varepsilon_t,
        \ldots)& \ \sim \mathcal N(0,\sigma^2)^{
        \otimes \mathbb N^*}.
    \end{matrix}
\end{align}
Here, for a probability distribution $Q$, we denoted by 
$Q^{\otimes \mathbb N^*}$ the probability distribution of the
sequence $(Z_i)_{i\in\mathbb N^*}$ of independent random
variables drawn from the same distribution $Q$.    
We denote the contextual best reward for arm $i$ by 
\[
    f_i^*(\mathbf{c}) \;=\; \sup_{\mathbf{x}\in\mathfrak{X}_{
    \mathbf{X}(i)}} f_i(\mathbf{x}, \mathbf{c}),
\]
and the corresponding contextual best action by 
\[
    \mathbf{x}^*_i(\mathbf{c}) \;=\; 
    \arg\max_{\mathbf{x}\in\mathfrak{X}_{\mathbf{X}(i)}} 
    f_i(\mathbf{x}, \mathbf{c}),
\]
which we refer to as the optimal policy of arm $i$. 

Throughout this section, we will use the notation
$\mathbb{P}(Z \mid \boldsymbol{f})$ and $\mathbb{E}[Z 
\mid \boldsymbol{f}]$ to denote, respectively, the 
conditional probability and the conditional expectation 
of a random vector $Z$ given $\boldsymbol{f} = (f_1, 
\ldots,f_m)$. The conditioning is understood with 
respect to the Gaussian process $\boldsymbol{f} =
(f_1,\ldots,f_m)$, which means that the integration 
is carried out over all realisations of the noise 
variables $\varepsilon_t$.

An agent, without prior knowledge of the functions $f_i$, interacts 
with this bandit in a sequential manner. At each time step $t$:
\begin{enumerate}
    \item the agent chooses $i_t \in [m]$,
    \item the agent observes $\mathbf{c}_{t} \sim P_{i_t}$,
    \item the agent selects $\mathbf{x}_{t} \in \mathfrak{X}_{
    \mathbf{X}(i_t)}$,
    \item the agent observes $y_t = f_{i_t}(\mathbf{x}_{t}, 
    \mathbf{c}_{t}) + \varepsilon_t$, where $\varepsilon_t 
    \overset{iid}{\sim} \mathcal{N}(0,\sigma^2)$.
\end{enumerate}
We use $\mathbf{x}_{i,s}$ and $\mathbf{c}_{i,s}$ to refer to the 
interventional and contextual values of the $i^{\text{th}}$ 
equivalence class of POMPSs when it has been selected for the 
$s^{\text{th}}$ time. Additionally, we refer to an equivalence 
class as an \emph{arm}, reflecting the similarity of our problem 
to the $m$-armed bandit setting. Each arm $i$ has $f_i$ as its 
expected payout function, where $\mathfrak{X}_{\mathbf{X}(i)}$ is 
assumed to be convex and compact. 

In our causal optimisation framework, contexts for each 
arm are sampled i.i.d.\ from a fixed distribution. A key 
aspect of our setting is that the agent actively selects 
which contextual variables to observe, a decision 
determined by the chosen POMPS. The agent's objective 
is to maximize (in expectation) the cumulative reward 
$\sum_{t=1}^T y_t$ over an unknown time horizon $T$.

Our proposed algorithm relies on the upper confidence bound 
(UCB) as the selection criterion for the evaluation point. 
Recall that the 
posterior mean and variance of the GP are denoted by $\mu_{i,s}
(\mathbf{x}, \mathbf{c})$ and $\sigma_{i,s}(\mathbf{x}, 
\mathbf{c})$, respectively, after the $i^{\text{th}}$ arm has 
been played $s$ times. The UCB at the point $(\mathbf{x}, 
\mathbf{c})$ after $s$ plays is defined as
\begin{align}\label{eq:UCB}
    \UCB_{i,\ell}(\mathbf{x}, \mathbf{c}) 
    = \mu_{i,\ell-1}(\mathbf{x}, \mathbf{c}) 
    + \sqrt{\beta_i(\ell)}\,\sigma_{i,\ell-1}
    (\mathbf{x}, \mathbf{c}),
\end{align}
for some tuning parameters $\beta_i(\ell)$. We 
summarise our method in \cref{alg:coca_bo}, where 
$\beta_{i,s}$ and $\pi_s$ are free parameters to be 
specified later.

\begin{algorithm}[t]
\caption{Intervention selection algorithm UCB-BO}
\label{alg:coca_bo}
\begin{algorithmic}
   \STATE {\bfseries Input:} $m:= \text{number of 
   equivalence classes of MPSs from }\mathbb{S}^*$ 
   \STATE {\bfseries Parameters:} $(\beta_i(n))_{i\in 
   [m];n\in\mathbb N^*}$, $(\rho_i(n))_{i\in [m]; 
   n\in \mathbb N^*}$.
   \FOR{$t=1$ {\bfseries to} ...}
   \IF{$t\leq m$}
   \STATE Select an equivalence class: $i_t:=t$.
   \STATE Observe context $\mathbf c_t = \mathbf{c}_{i_t, 1}$.
   \STATE Select an action: $\mathbf x_t = \mathbf{x}_{i_t,1} = \arg\max_{ 
   \mathbf{x} \in\mathfrak{X}_{\mathbf{X}(i_t)}} \UCB_{i_t,1} 
   (\mathbf{x}, \mathbf{c}_{i_t, 1})$. {(See Eq. \eqref{eq:UCB} for UCB.)}
   \STATE Set: $\bar{U}_{i_t}(1) := \UCB_{i_t,1}(\mathbf{x}_{i_t,1},\mathbf{c}_{i_t, 1}); \ n_{i_t}=1$.
   \STATE Observe: $y_t$
   \STATE Update the posterior for the $i_t$ Gaussian 
   process based on $\mathbf{x}_{i_t,1},\mathbf{c}_{
   i_t, 1},\text{ and } y_t$.
   \ELSE
   \STATE Select an equivalence class: $i_t= \arg
   \max_{i\in [m]}  \big( \bar{U}_{i}(n_i)+\frac{\rho_i 
   (n_i)}{\sqrt{n_i}}\big)$.
   \STATE Update: $n_{i_t}=n_{i_t}+1$.
   \STATE Observe context: $\mathbf c_t=\mathbf{c}_{i_t, n_{i_t}}$
   \STATE Select an action: $\mathbf x_t = \mathbf{x}_{i_t, n_{i_t}}=\arg\max_{\mathbf{x}\in\mathfrak{X}_{\mathbf{X}_{i_t}}} 
   \UCB_{i_t, n_{i_t}}(\mathbf{x}, \mathbf{c}_{t})$
   \STATE Observe: $y_t$
   \STATE Update the posterior for the $i_t$ Gaussian process based on $\mathbf{x}_t,\ \mathbf{c}_t, 
   \text{ and } y_t$.
   \STATE Set: $\bar{U}_{i_t}(n_{i_t}) = \frac{(n_{i_t}-1)
   }{n_{i_t}} \bar{U}_{i_t}(n_{i_t}-1) + \frac1{n_{i_t}} 
   \UCB_{i_t,n_{i_t}}(\mathbf x_t, \mathbf c_t)$
   \ENDIF
   \ENDFOR
\end{algorithmic}
\end{algorithm}

\paragraph{Regret Definition}

Let $\mu_i=\mathbb{E}[f_i^*(\mathbf{C}_{i,1})\,|\,f_i] = \int \max_{\mathbf x} f_i(\mathbf x,\mathbf c)\,P_i(
\mathrm d\mathbf c)$ and $I^* = \argmax_{i\in[m]} \mu_i$. 
Clearly,  $\mu_i$ depends on $f_i$, whereas $R_T$ and $I^*$ 
depend on $\boldsymbol{f}$, but for the sake of simplicity 
of writing, this dependence will not be highlighted in the 
notation. We also write $\mu^* = \max_{i\in[m]} \mu_i$. 
The regret then is defined as:
\begin{align}\label{eq:reg_def}
    R_T& = \sum_{t=1}^T \big\{\mu^* - \mathbb E[f_{i_t}
    (\mathbf{X}_{t}, \mathbf{C}_t)\mid \boldsymbol{f} 
    ,\mathcal D^{t-1}]\big\}.
\end{align}
The regret $R_T$ can be decomposed into two components:
\begin{align}
    R_T &= \sum_{t=1}^T\Big\{\underbrace{
    \mu^* - \mu_{i_t}}_{\text{Arm 
    Selection Regret}} + \underbrace{\mu_{i_t} - 
    \mathbb E[f_{i_t}(\mathbf{X}_{t}, \mathbf{C}_t)\mid 
    \boldsymbol{f},\mathcal D^{t-1}]}_{
    \text{Context-averaged Action Selection Regret}}\Big\},
\end{align} 
The context-averaged action selection regret measures how 
much the mean reward $f_{i_t}$ of the action $\mathbf{x}_t$ 
chosen by the agent deviates from the mean reward of the 
best possible action $\mathbf{x}^*_{i}(\mathbf{c}_{i_t})$, 
averaged over all values of the context. 
We note that it is localised to the chosen arm. The arm 
selection regret compares the rewards of each arm if the agent 
plays according to the optimal policy of that arm.

\subsection{Proof of the main result}

Our next lemmas motivate the choice of the index for arms 
in \Cref{alg:coca_bo}.  Recall that $\mathbf X_{i,\ell}$ is 
an element of $\mathfrak{X}_{\mathbf{X}(i)}$ that maximises
$\mathbf x\mapsto \textrm{UCB}_{i,\ell}(\mathbf x,\mathbf 
C_{i,\ell})$.

\begin{lemma}\label{lem:event}
    Let $\delta\in(0,1)$ and $(\pi_n)_{n\in\mathbb N^*}$ 
    be a nondecreasing sequence of positive numbers such 
    that $\sum_{n\in\mathbb N^*} \pi_n^{-1} = 1$. Let 
    $\mathcal N_{i,n}$ be an $\varepsilon$-net of $\mathfrak{
    X}_{\mathbf{X}(i)}$ in $\ell_1$-norm, with $\varepsilon 
    = 1/(m\pi_n r_i d_i)$. If for all $n\in\mathbb N^*$,
    \begin{align}
        \beta_{i}(n) &\geq  
        2 \log(2/(m\delta)) + 4d_i \log_+(mr_id_i \pi_n),
    \end{align}
    there is an event $\mathcal A_i(\delta)$ of probability 
    at least $1-\delta$ such that on $\mathcal A_i(\delta)$,
    the following properties hold
    \begin{align}
        f_i(\mathbf X_{i,\ell},\mathbf C_{i,\ell}) &\ge
        \textrm{\rm UCB}_{i,
        \ell}(\mathbf X_{i,\ell}, \mathbf C_{ i,\ell}) -
        2\sqrt{\beta_i(\ell)}\, \sigma_{i,\ell-1}( 
        \mathbf X_{i,\ell}, \mathbf C_{i, \ell});\text{ 
        \rm for all } \ell\in\mathbb N^*,\label{eq:16}\\
        f_i^*(\mathbf C_{i,\ell}) &\leq \textrm{
        \rm UCB}_{i,\ell}(\mathbf X_{i,\ell}, \mathbf 
        C_{i,\ell}) + \frac{\varphi}{m\pi_\ell} 
        \sqrt{\log(2\smash{\bar d_i}\psi /\delta)};  \quad \text{\rm 
        for all}\ \mathbf x \in \mathfrak{X}_{\mathbf{X}(i)};\ 
        \ell\in\mathbb N^*.\label{eq:17}
    \end{align}
\end{lemma}
\begin{proof}
    We define the events
    \begin{align}
        \Omega_0 & = \big\{|f_i(\mathbf{x}, \mathbf{c})-f_i(
        \mathbf{x}', \mathbf{c})|\leq \varphi \sqrt{\log({ 
        2\smash{\bar d_i}\psi}/{\delta})}\,\|\mathbf{x}-\mathbf{x}'\|_1,
        \ \text{ for all }\ \mathbf{x},\mathbf{x}'\in 
        \mathfrak{X}_{\mathbf{X}(i)},\ \mathbf{c}
        \in \mathfrak{X}_{\mathbf{C}(i)}\big\},\\[5pt]
        \Omega_{\ell,1} & = \big\{f_i(\mathbf x,\mathbf 
        C_{i,\ell}) \leq \textrm{UCB}_{i,\ell}(\mathbf x,
        \mathbf C_{i,\ell});\quad \text{for all }\ \mathbf x
        \in\mathcal N_{i,\ell} \big\},\\[5pt]
        \Omega_{\ell,2} & = \Big\{ f_i^*(\mathbf C_{i,\ell}) 
        \ge \textrm{UCB}_{i,\ell}(\mathbf X_{i,\ell}, 
        \mathbf C_{ i,\ell})- 2\sqrt{\beta_i(\ell)}\,
        \sigma_{i,\ell-1}(\mathbf X_{i,\ell}, \mathbf C_{i, 
        \ell})\Big\}.
    \end{align}
    \Cref{ass:2} implies that $\mathbb P(\Omega_0) 
    \geq 1-\delta/2$. Furthermore, we recall that for any
    $(\mathbf x,\mathbf c) \in\mathfrak{X}_{\mathbf{X}(i)} 
    \times\mathfrak{X}_{\mathbf C(i)}$ the conditional
    distribution of $f_i(\mathbf x, \mathbf c)$ given 
    $\mathcal D_{i,\ell-1}$ is Gaussian with mean 
    $\mu_{i,\ell-1}(\mathbf x,\mathbf c)$ and variance 
    $\sigma_{i,\ell-1}^2(\mathbf x,\mathbf c)$. This implies 
    that
    \begin{align}
        \mathbb P\Big( f_i(\mathbf x,\mathbf c) \leq 
        \mu_{i,\ell-1}(\mathbf x,\mathbf c) + \sqrt{2\log(1/
        \delta_\ell)} \, \sigma_{i,\ell-1}(\mathbf x,\mathbf 
        c) \,\Big|\,\mathcal D_{i,\ell-1}\Big) \geq 1 - \delta_\ell/2,
    \end{align}
    for every $\delta_\ell\in(0,1)$. Note that the above 
    inequality holds not only for any deterministic $(\mathbf 
    x,\mathbf c)$, but also for every random pair $(\mathbf x, 
    \mathbf c)$ provided that it is independent of $f_i$ given
    $\mathcal{D}_{i,\ell-1}$. Therefore, applying this 
    inequality to all pairs $\{(\mathbf x,\mathbf C_{i,\ell}):
    \mathbf x\in \mathcal N_{i,\ell}\}$ and taking the union 
    bound, we get
    \begin{align}
        \mathbb P\Big( f_i(\mathbf x,\mathbf C_{i,\ell}) \leq 
        \mu_{i,\ell-1}(\mathbf x,\mathbf C_{i,\ell}) + 
        \sqrt{\beta_i(\ell)} \, \sigma_{i,\ell-1}(\mathbf x,
        \mathbf C_{i,\ell});\ \forall\mathbf x\in 
        \mathcal N_{i,\ell} \,\Big|\,\mathcal D_{i,\ell-1}
        \Big) \geq 1 - \frac{|\mathcal N_{i,\ell}|
        \delta_\ell}2, 
    \end{align}
    as soon as $\beta_i(\ell)\geq 2\log(1/\delta_\ell)$. 
    Taking into account the definition of $\textrm{UCB}_{i, 
    \ell}$, we arrive at $\mathbb P(\Omega_{\ell,1}| 
    \mathcal D_{i,\ell-1} ) \geq 1-|\mathcal N_{i,\ell}| 
    \delta_\ell/2$. Since this inequality holds for any set 
    of values taken by the data $\mathcal D_{i,\ell-1}$, we 
    get 
    \begin{align}
        \mathbb P(\Omega_{\ell,1}) \geq 1- (\delta/4\pi_\ell)
        , \quad\text{provided that}\quad \beta_i(\ell)\geq 
        2\log(2\pi_\ell |\mathcal N_{i,\ell}|/\delta). 
    \end{align} 
    Since $\mathbf X_{i,\ell}$ is the maximiser in 
    $\mathfrak{X}_{\mathbf{X}(i)}$ of $\mathbf x\mapsto 
    \textrm{UCB}_{i,\ell}(\mathbf x,\mathbf C_{i,\ell})$, 
    it is clearly independent of $f_i$ conditionally to 
    $\mathcal D_{i, \ell-1}$. Therefore, the argument used 
    above can be repeated  to infer that the following 
    inequalities hold with probability $\geq 1 - \delta/
    (4\pi_\ell)$:
    \begin{align}
        f_i (\mathbf X_{i,\ell}, \mathbf C_{i,\ell})
        & \geq \mu_{i,\ell-1}(\mathbf X_{i,\ell}, \mathbf 
        C_{i,\ell}) - \sqrt{2\log(2\pi_\ell/\delta)}\,
        \sigma_{i,\ell-1}(\mathbf X_{i,\ell}, \mathbf 
        C_{i,\ell})\nonumber\\
        & \geq \textrm{UCB}(\mathbf X_{i,\ell}, 
        \mathbf C_{i,\ell}) -  2\sqrt{\beta_i(\ell)}\,
        \sigma_{i,\ell-1}(\mathbf X_{i,\ell}, \mathbf 
        C_{i,\ell}),
    \end{align}
    provided that $\beta_i(\ell)\geq 2\log(2\pi_\ell/
    \delta)$. We conclude that as soon as $\beta_i(\ell) 
    \geq 2\log(2\pi_\ell |\mathcal N_{i,\ell}
    |/\delta)$, we have $\mathbb P(\mathcal A_i(\delta)) 
    \ge 1-\delta$ for 
    \begin{align}
        \mathcal A_i(\delta) = \Omega_0\cap \Big\{\bigcap_{
        \ell=1}^\infty \big(\Omega_{\ell,1}\cap\Omega_{
        \ell,2}\big)\Big\}. 
    \end{align}
    The fact that \eqref{eq:16} holds on $\mathcal{A} 
    (\delta)$ is obvious. To show that \eqref{eq:17} 
    holds as well, it suffices to notice that for all 
    $\mathbf x \in \mathfrak{X}_{\mathbf{X}(i)}$ and for 
    all $\ell\in\mathbb N^*$,
    \begin{align}
        f_i(\mathbf x,\mathbf C_{i,\ell}) &\leq 
        \max_{\mathbf x\in\mathcal N_{i,\ell}} 
        f_i(\mathbf x,\mathbf C_{i,\ell}) + \varphi 
        d_ir_i\sqrt{\log(2\smash{\bar d_i}\psi/\delta)}\,(m\pi_\ell
        r_id_i)^{-1}\\
        &\le \textrm{UCB}_{i,\ell}(\mathbf x, \mathbf 
        C_{i,\ell}) + \varphi \sqrt{\log(2d_i
        \psi /\delta)}\,(m\pi_\ell)^{-1}\\
        &\le \textrm{UCB}_{i,\ell}(\mathbf X_{i,\ell}, 
        \mathbf C_{i,\ell}) + \varphi \sqrt{
        \log(2\smash{\bar d_i}\psi /\delta)}\,(m\pi_\ell)^{-1}.
    \end{align}
    The first inequality above follows from the fact that
    $\Omega_0$ is realised, the second inequality follows
    from $\Omega_{\ell,1}$ and the third inequality is due
    to the fact that $\mathbf X_{i,\ell}$ is the maximiser
    of $\mathbf x\mapsto\text{UCB}_{i,\ell}(\mathbf x, 
    \mathbf C_{i,\ell})$. 
    Since we assumed that $\mathfrak{X}_{\mathbf{
    X}(i)}\subset [0,r_i]^{d_i}$, we have $|\mathcal 
    N_{i,\ell}|\leq (r_id_i/\varepsilon)^{d_i}\vee 1 =  (mr_i^2d_i^2\pi_\ell)^{d_i}\vee 1$. This
    implies that it suffices to choose $\beta_{i}(\ell)$
    satisfying
    \begin{align}
        \beta_{i}(\ell) \geq 2 \log(2/(m\delta)) + 
        4d_i \log_+(m r_id_i \pi_\ell). 
    \end{align}
    This completes the proof of the lemma.
\end{proof}

\begin{lemma}\label{lem:bound_on_smp_mean}
    Let $(\pi_n)_{n\in\mathbb N^*}$ be a nondecreasing 
    sequence of positive numbers such that $\sum_{n\in
    \mathbb N^*} \pi_n^{-1} = 1$. Define
    \begin{align}
        \bar U_i(n)  = \frac1n\sum_{\ell=1}^n 
        \text{\rm UCB}_{i,\ell} (\mathbf{X}_{i,\ell}, 
        \mathbf{C}_{i,\ell}),\qquad
        \bar{\mu}_i(n) =\frac{1}{n}\sum_{\ell=1}^n 
        f^*_{i}(\mathbf{C}_{i,\ell}).
    \end{align}
    If \eqref{eq:16} and \eqref{eq:17} are satisfied for
    some $\delta \in(0,1)$, then for every $n\in\mathbb N^*$,
    \begin{align}\label{eq:62}
        \bar{\mu}_i(n) - \frac{\varphi \sqrt{\log(2d_i
        \psi/\delta)}}{mn} 
        \leq\bar{U}_i(n) \leq \bar\mu_i(n) + 2\sqrt{\frac{2
        \kappa_i\beta_{i}(n)\gamma_{i}(n)}{n\log(1 + 
        \smash{\sigma^{-2}\kappa_i})}}.
    \end{align}
\end{lemma}
\begin{proof}
    From \eqref{eq:17}, we infer that
    \begin{align}
        f_i^*(\mathbf C_{i,\ell}) & \le 
        \textrm{UCB}_{i,\ell}(\mathbf X_{i,\ell}, \mathbf
        C_{i,\ell}) + \frac{\varphi}{m\pi_\ell}\sqrt{ 
        \log(2\smash{\bar d_i}\psi/\delta)}. 
    \end{align}
    Averaging over $\ell=1,\ldots,n$, and using the 
    inequality $\sum_{\ell=1}^n 1/\pi_\ell \le 1$, we obtain
    \begin{align}
        \bar\mu_i(n) \le \bar U_i(n) + \frac{\varphi d_ir_i
        }{mn}\sqrt{\log(2\smash{\bar d_i}\psi/\delta)}. 
    \end{align}
    To get a lower bound on $\bar\mu_i(n)$, we average over $\ell = 1,\ldots,n$ inequalities \eqref{eq:16}, 
    then apply the Cauchy-Schwarz inequality, to obtain
    \begin{align}
        \bar\mu_i(n) &\ge \bar U_i(n) - \frac2n
        \sum_{\ell=1}^n \sqrt{\beta_i(\ell)}\,
        \sigma_{i,\ell-1}(\mathbf X_{i,\ell},\mathbf 
        C_{i,\ell})\\
        & \ge \bar U_i(n) - \frac{2\sqrt{\beta_{i}(n)}
        }{n} \sum_{\ell=1}^n \sigma_{i,\ell-1}(\mathbf 
        X_{i,\ell},\mathbf C_{i,\ell})\\
        &\ge \bar U_i(n) - \frac{2\sqrt{\beta_{i}(n)}
        }{\sqrt{n}} \bigg\{\sum_{\ell=1}^n 
        \sigma_{i,\ell-1}^2(\mathbf X_{i,\ell},
        \mathbf C_{i,\ell})\bigg\}^{1/2}.\label{eq:65}
    \end{align}
    The definition of $\sigma^2_{i,\ell}(\mathbf x,\mathbf 
    c)$ as the posterior variance of $f_i(\mathbf x,\mathbf 
    c)$ given $\mathcal D_{i,\ell-1}$ and formula 
    \eqref{sigma_post} imply that $\sigma^2_{i,\ell}
    (\mathbf x,\mathbf c)\leq \kappa_i$ for every $(
    \mathbf x,\mathbf c)$. Therefore, the increasingness
    of the function $x\mapsto x/\log(1+x)$ yields
    \begin{align}
        \frac{\sigma^{-2}\sigma_{i,\ell-1}^2(\mathbf x,
        \mathbf C_{i,\ell})}{\log\big(1 + \sigma^{-2}
        \sigma_{i,\ell-1}^2(\mathbf x, \mathbf C_{i,\ell})
        \big)} \leq \frac{\sigma^{-2}\kappa_i}{\log(1 + 
        \sigma^{-2}\kappa_i)}.
    \end{align}
    This can be rewritten as 
    \begin{align}
        \sigma_{i,\ell-1}^2(\mathbf X_{i,\ell},
        \mathbf C_{i,\ell}) \leq \frac{\kappa_i \log\big(1 
        + \sigma^{-2} \sigma_{i,\ell-1}^2(\mathbf X_{i,\ell}, 
        \mathbf C_{i,\ell}) \big)}{\log(1 + \sigma^{-2}
        \kappa_i)},\qquad \forall \ell\in\mathbb N^*.
    \end{align}
    Therefore, we have 
    \begin{align}
        \sum_{\ell=1}^n \sigma_{i,\ell-1}^2(\mathbf X_{i,\ell},
        \mathbf C_{i,\ell}) &\le \frac{\kappa_i}{\log(1 + 
        \sigma^{-2}\kappa_i)} \sum_{\ell=1}^n\log\big(1 
        + \sigma^{-2} \sigma_{i,\ell-1}^2(\mathbf X_{i,\ell}, 
        \mathbf C_{i,\ell}) \big)\\
       &\le \frac{2\kappa_i\gamma_{i}(n)}{\log(1 + 
        \sigma^{-2}\kappa_i)}.\label{eq:66}
    \end{align}
    The last inequality above follows from the definition
    of $\gamma_n$ and \citep[Lemma 5.3]{srinvas_kauss}. 

    Combining \eqref{eq:65} and \eqref{eq:66}, we arrive at
    \begin{align}
        \bar U_i(n) \le \bar\mu_i(n) + 2\sqrt{\frac{2
        \kappa_i\beta_{i}(n)\gamma_{i}(n)}{n\log(1 + 
        \sigma^{-2}\kappa_i)}}. 
    \end{align}
    This completes the proof of the lemma.
\end{proof}

\begin{lemma}\label{lem:Hoeffding}
    Let $\delta\in(0,1)$ and let $\mathfrak{X}_{\mathbf C(i)}$
    be convex and compact, included in $[0,r_i']^{\dprimei }$, 
    for some $\dprimei \in\mathbb N^*$ and some $r_i'\ge 0$. Let
    us denote 
    \begin{align}\label{eq:lambda'}
        \lambda_i'(\delta) = \dprimei r_i'\varphi\log(2
        (\dprimei \psi+2\pi_n)/\delta).
    \end{align}
    There exists an event $\mathcal A'_i(\delta)$ of 
    probability at least $1-\delta$ such that on 
    $\mathcal A'_i(\delta)$, it holds
    \begin{align}
        \bigg|\frac1n\sum_{\ell=1}^n f_i^*(\mathbf
        C_{i,\ell}) - \mu_i\bigg|\le \frac{\lambda_i'(\delta)
        }{\sqrt{2n}},\qquad\forall n\in\mathbb N^*.
    \end{align}
\end{lemma}
\begin{proof}
    If we set $\Delta f_i^* = \max_{\mathbf c} f_i^*(
    \mathbf c) - \min_{\mathbf c} f_i^*(\mathbf c)$, the H\"offding inequality implies that  
    \begin{align}
        &\mathbb P \bigg(\bigg|\frac1n\sum_{\ell=1}^n 
        f^*_i(\mathbf{C}_{i,\ell}) - \mu_i\bigg| 
        \leq \Delta f_i^*\,\sqrt{\frac{\log(2/\delta_1)}{2n}}\,
        \Big|\, \boldsymbol{f}\,\bigg)  
        \ge 1-\delta_1,\quad\forall i\in [m].  
    \end{align}
    Choosing $\delta_1 = \delta/(2\pi_n)$ and using the 
    Bonferroni inequality, we get 
    \begin{align}\label{eq:63}
        \mathbb P \bigg(\bigg|\frac1n\sum_{\ell=1}^n 
        f^*_i (\mathbf{C}_{i,\ell}) - \mu_i\bigg| \leq 
        \Delta f_i^*\,\sqrt{\frac{\log(4\pi_n/\delta)}{2n}}
        \,\text{ for all }n\in\mathbb N^*\,
        \Big|\, \boldsymbol{f}\,\bigg) \ge 1-\delta/2.  
    \end{align}
    On the other hand, since $\mathfrak{X}_{\mathbf C(i)}$
    is convex and compact, included in $[0,r_i']^{\dprimei }$, 
    \Cref{ass:2} implies that there exist kernel dependent
    positive constants $\varphi$, $\psi$ such that,
    with probability at least $1-\delta/2$,
    \begin{align}
        |f_i(\mathbf x,\mathbf c) - f_i(\mathbf x, 
        \mathbf c')|\le \varphi \sqrt{\log(2 \smash{\dprimei  
        \psi/\delta})}\,\|\mathbf c - 
        \mathbf c'\|_1;\quad \text{for all }
        \mathbf x\in\mathfrak{X}_{\mathbf{X}(i)}, \mathbf c,
        \mathbf c'\in\mathfrak{X}_{\mathbb C(i)}.
    \end{align}
    Hence, with probability at least $1-\delta/2$, we have 
    \begin{align}
        \Delta f_i^*\leq \dprimei  r_i'.\varphi \sqrt{
        \log(2 \smash{\dprimei  \psi/\delta})}. 
    \end{align}
    Combining with \eqref{eq:63}, this implies that on an 
    event of probability at least $1-\delta$, denoted by 
    $\mathcal A_i'(\delta)$, we have, for any $n\in\mathbb N^*$, 
    \begin{align}
        \bigg|\frac1n\sum_{\ell=1}^n f^*_i
        (\mathbf{C}_{i,\ell}) - \mu_i\bigg| &\leq  
        \dprimei  r_i'\varphi \sqrt{
        \log(2 \smash{\dprimei  \psi/\delta})}
        \,\sqrt{\frac{\log(4\pi_n/\delta)}{2n}}\\
        &\le \dprimei  r_i'\varphi \,\frac{\log(2(\dprimei \psi
        +2\pi_n)/\delta)}{\sqrt{2n}}=\frac{\lambda_i'
        (\delta)}{\sqrt{2n}}.
    \end{align}
    This completes the proof of the lemma. 
\end{proof}
Our next step is to upper bound the number of times a 
suboptimal arm $i\not\in I^*$ is played.

\begin{lemma}\label{lem:sub_opt_play}
    Let $\delta\in(0,1/2)$ and $(\pi_n)$ be a positive 
    nondecreasing sequence satisfying $\sum_{n\in 
    \mathbb N^*} 1/\pi_n = 1$. Let $\mathsf C_i = 
    8\kappa_i/\log(1+\sigma^{-2}\kappa_i)$. Assume 
    that the parameters $\beta_i(n)$ and $\rho_i(n)$
    are selected so that, for every $i\in[m]$ and 
    every $n\in\mathbb N^*$,
    \begin{align}
        \beta_{i}(n) &\geq  2 \log(4/\delta) + 4d_i
        \log(mr_id_i\pi_n),\\
        \rho_i(n) &\ge {\varphi}\sqrt{
        \log(4m\smash{\bar d_i}\psi /\delta)} +  \dprimei r_i'\varphi 
        \log\big(4m(\dprimei \psi+2\pi_n)/\delta\big).
    \end{align}
    Let $\mathcal A(\delta) = \bigcap_{i=1}^m \big(\mathcal A_i
    (\delta/2m)\cap \mathcal A_i'
    (\delta/2m)\big)$ be the intersection of the events
    defined in \Cref{lem:event} and \Cref{lem:Hoeffding}. 
    Then $\mathbb P(\mathcal A(\delta))\ge 1-\delta$ and 
    on $\mathcal A(\delta)$, the number of times 
    $n_i=n_i(T)$ a suboptimal arm with a gap $\Delta_i>0$ 
    has been played over the horizon $T$ satisfies
    \begin{align}\label{eq:action_bound_disc}
        \sqrt{n_i(T)} \le \frac{\sqrt{\mathsf C_i 
        \beta_i(n_i) \gamma_{i} (n_i)} + 2 
        \rho_i(n_i)}{\Delta_i}\bigvee 1, 
        \quad \forall T\geq 1,\,\forall i\not\in  I^*.
    \end{align}
\end{lemma}

\begin{proof}
The fact that $\mathbb P\big(\mathcal A(\delta)\big)
\ge 1-\delta$ follows from \Cref{lem:event}, 
\Cref{lem:Hoeffding} and the union bound. 
Let us use the notations
\begin{align}
    b_i(n) = \sqrt{\mathsf C_i\beta_{i}(n)\gamma_{i}
    (n)} 
\end{align}
and $\lambda_i'$ as defined by \eqref{eq:lambda'}. 
For all pairs $(T,i)$ such that $n_i(T)= 1$, the desired
inequality is obviously satisfied. Consider now the pairs
$(T,i)$ such that $n_i(T)\ge 2$. 
Let $t_i$ be the last time the arm $i$ has been played
among the first $T$ rounds. To ease notation, let us set 
$n_i = n_i(t_i) = n_i(T)$. Let $j = j(\boldsymbol{f})$ 
be an optimal arm, that is, an integer from $[m]$ 
maximising $i\mapsto \mu_i$. Since at round $t_i$ arm $i$ 
has been preferred to the optimal arm $j$, we have
\begin{align}\label{eq:19}
    \bar{U}_i(n_i) + \frac{\rho_i(n_i)}{\sqrt{{n_i}}} 
    \geq \bar{U}_j(n_j) + \frac{\rho_j(n_j)}{\sqrt{{n_j}}}.
\end{align}
Under our choice of $\beta_{i}(n)$ and $\pi_n$, 
\Cref{lem:event} and \Cref{lem:bound_on_smp_mean} 
applied with $\delta/(2m)$ instead of $\delta$, 
imply that on $\mathcal A(\delta)$, 
for all $T\geq 1$ and for all $i\not\in I^*$, it holds 
that
\begin{align}
    \bar U_i(n_i) &\leq \bar{\mu}_i(n_i) + \frac{b_i
    (n_i)}{\sqrt{n_i}},\label{eq:20a}\\
    \bar U_j(n_j)&\geq \bar{\mu}_j(n_j) -  
    \frac{\varphi_j \sqrt{\log(4md_j\psi_j 
    /\delta)}}{mn_j}.\label{eq:20b}
\end{align}
Combining \eqref{eq:19}, \eqref{eq:20a} and \eqref{eq:20b}, 
we check that with probability at least $1-\delta/2$, for 
all $T\geq 1$ and for all $i\not\in I^*$, it holds that
\begin{align}
    \bar{\mu}_i(n_i) + \frac{b_i(n_i) + \rho_i(n_i)}{
    \sqrt{{n_i}}} \geq \bar{\mu}_j(n_j) + \frac{\rho_j
    (n_j)}{\sqrt{\smash{n_j}\vphantom{n}}} - \frac{\varphi_j \sqrt{\log(4md_j\psi_j /\delta)}}{mn_j} .\label{eq:21}    
\end{align}
Since we work under $\mathcal{A}(\delta)$ that
is included in $\bigcup_{i=1}^m \mathcal A_i'(\delta/2m)$, 
we have
\begin{align}
    |\bar\mu_i(n) - \mu_i| \leq \frac{\lambda_i'(\delta/2m)}
    {\sqrt{2n}},\qquad \text{for all } i\in[m].
\end{align}
In conjunction with \eqref{eq:21}, this implies that 
for all $T\geq 1$ and for all $i\not\in I^*$, $j\in I^*$, 
\begin{align}
    \mu_i + \frac{b_i(n_i) + \rho_i(n_i)}{\sqrt{n_i}} 
    + \frac{\lambda_i'(\delta/2m)}{\sqrt{2n_i}} \geq \mu_j 
    + \frac{\rho_j(n_j)}{\sqrt{\smash{n_j}\vphantom{n}}} 
    - \frac{\varphi_j \sqrt{\log(4md_j\psi_j/\delta)}}{mn_j}
    - \frac{\lambda_j'(\delta/2m)}{\sqrt{2n_j}}.
    \label{eq:22}    
\end{align}
In the rest of this proof, we work under this event. 

When $\rho_j(n)$ is chosen so that  for all $j\in[m]$, 
\begin{align}
    \rho_j(n) \ge \frac{\varphi_j}{m} \sqrt{\frac{\log(4m\smash{d_j\psi_j/\delta)}}{n}} 
    + \frac{\lambda_j'(\delta/2m)}{\sqrt{2}},
\end{align}
we infer that
\begin{align}
    \frac{b_i(n_i) + \rho_i(n_i) + \lambda_i'(\delta/2m)}
    {\sqrt{\smash{n_i}\vphantom{n}}} \geq \Delta_i,
    \qquad \text{for all }i\not\in I^*.
\end{align}
This implies that 
\begin{align}
    \sqrt{n_i}&\le \frac{b_i(n_i) + \rho_i(n_i) + 
    \lambda'_i(\delta/2m)}{\Delta_i}
    \le \frac{b_i(n_i) + 2\rho_i(n_i)}{
    \Delta_i}.
\end{align}
This completes the proof of the lemma.
\end{proof}

From now on, we use the notation
\begin{align}
    \mathsf C &= \max_i \mathsf C_i,\quad 
    \kappa = \max_i \kappa_i,\quad
    d = \max_i d_i,\quad d' = \max_i \dprimei ,\\
    \beta(T) &= \max_{i}\max_{n\le T} \beta_i(n),\quad
    \rho(T) = \max_{i}\max_{n\le T} \rho_i(n),
\end{align}
and
\begin{align}
    \gamma(T) = \max_{i} \gamma_i(T),\quad
    \bar\gamma (T) = \max_{(n_i): \sum_i n_i = T}
    \frac1m \sum_{i=1}^m \gamma_i(n_i), 
\end{align}
the last maximum being over all the sequences of 
positive integers summing to $T$. We omit the 
dependence of $n_i(T)$ on the horizon $T$ and 
write $n_i$ for brevity. We denote $\beta_I = 
\max _{i\in I}\beta_{i,T}$ and upper bound $R_T$ 
under \Cref{lem:sub_opt_play}.

\begin{proposition}\label{prop:reg_bound}
    Let $\delta\in(0,1/2)$ and $(\pi_n)$ be a positive 
    nondecreasing sequence satisfying $\sum_{n\in\mathbb N^*} 
    1/\pi_n = 1$. Let $\mathsf C = 8\kappa/\log(1 + 
    \sigma^{-2}\kappa)$.
    Let $\boldsymbol{\Delta} = (\Delta_1,\ldots,\Delta_m)$
    be the vector of sub-optimality gaps defined by 
    $\Delta_i = \mu^* - \mu_i$ and $I^*=\{i:\Delta_i=0\}$.    
    Let \Cref{ass:1} and 
    \Cref{ass:2} be fulfilled, and assume that the 
    parameters $\beta_{i}$ and $\rho_i$ of the 
    algorithm satisfy
    \begin{align}
            \beta_{i}(n) &\geq  2 \log(4/\delta) + 4d_i
        \log(mr_id_i\pi_n),\\
        \rho_i(n) &\ge {\varphi}\sqrt{
        \log(4m\smash{\bar d_i}\psi /\delta)} +  \dprimei r_i'\varphi 
        \log\big(4m(\dprimei \psi+2\pi_n)/\delta\big)
    \end{align}
    for every $i\in[m]$ and every $n\in\mathbb N^*$. 
    If the UCB-BO 
    \Cref{alg:coca_bo} is applied up to horizon $T$ then, 
    with probability at least $1-2\delta$, the regret defined 
    by \eqref{eq:reg_def} satisfies the following 
    two inequalities 
    \begin{align}\label{eq:reg_bound_lem}
        R_T &\leq 5\sqrt{mT}\,\big(\sqrt{\mathsf C
        \beta(T)\bar \gamma(T)} + \rho(T)\big) + 
        \|\boldsymbol{\Delta}\|_1 ,\\
        R_T &\le 2\sqrt{|I^*| T}\,\big(\sqrt{\mathsf 
        C \beta(T)\bar \gamma(T)} + \rho(T)\big) + 
        22\big(\mathsf C\beta(T) \gamma(T) + 
        \rho^2(T)\big)\sum_{i\not\in I^*} \frac{1}{
        \Delta_i}\\ 
        &\qquad + \|\boldsymbol{\Delta}\|_1 + 
        2m\big(\sqrt{\kappa\beta(1)} + \rho(T)\big).
    \end{align}
    for all $T\in\mathbb N^*$.
\end{proposition}

\color{black}
\begin{proof}
Throughout the proof, we assume that the event $\mathcal 
A(\delta)$, of probability at least $1-\delta$, defined 
in \Cref{lem:sub_opt_play}, is realised. We will also 
repeatedly use the following inequalities, that are 
direct consequences of the Cauchy-Schwarz inequality:
\begin{align}
    \sum_{i\in J} \sqrt{n_i(T)} \le \sqrt{|J|T},\qquad
    \sum_{i\in J} \sqrt{n_i(T)\rho_i(n_i(T))} \le 
    \sqrt{mT \bar\rho(T)},\quad
    \text{for all } J\subset [m].
\end{align}
We will also use the notation
\begin{align}
    c(n) = \sqrt{\mathsf C\beta(n)\gamma(n)} + \rho(n),\\
    \bar c(n) = \sqrt{\mathsf C\beta(n)\bar \gamma(n)} + 
    \rho(n),
\end{align}
for any $n\in\mathbb N^*$.
Let us define the functions $f_i^\circ(\mathbf x, 
\mathbf c) = f^*_i (\mathbf c) - f_i(\mathbf x, 
\mathbf c)$. With this notation, the regret $R_T$ can 
be written as:
\begin{align}
    R_T &= \sum_{t=1}^T\Big\{\mu^* - \mathbb E[f_{i_t}(
    \mathbf{X}_{t}, \mathbf{C}_t)\mid \boldsymbol{f},
    \mathcal D^{t-1}]\Big\}\\
    & = \sum_{t=1}^T\Big\{\mu^*-\mu_{i_t} 
    + \mathbb E[f_{i_t}^\circ(\mathbf X_t, \mathbf C_t)
    \mid \boldsymbol{f},\mathcal D^{t-1}]\Big\}\\
    & = \sum_{t=1}^T\{\mu^*-\mu_{i_t}\} + \sum_{t=1}^T 
    f_{i_t}^\circ(\mathbf X_t, \mathbf C_{t}) + 
    \sum_{t=1}^T \big\{\mathbb E [f_{i_t}^\circ(
    \mathbf X_t, \mathbf C_t) \mid \boldsymbol{f},
    \mathcal D^{t-1}] - f^\circ_{i_t} (\mathbf X_t,
    \mathbf C_t)\Big\}.
\end{align}
This leads to a decomposition of the regret into three
components
\begin{align}
    R_T = R_{T,1} + R_{T,2} + R_{T,3}
\end{align}
with 
\begin{align}
    R_{T,1}& = \sum_{t=1}^T\{\mu^*-\mu_{i_t}\},\qquad
    R_{T,2} = \sum_{t=1}^T f_{i_t}^\circ(\mathbf X_t, 
    \mathbf C_{t})\\
    R_{T,3}& = \sum_{t=1}^T \big\{\mathbb E [f_{i_t}^\circ(
    \mathbf X_t, \mathbf C_t) \mid \boldsymbol{f},
    \mathcal D^{t-1}] - f^\circ_{i_t} (\mathbf X_t,
    \mathbf C_t)\Big\}.
\end{align}
One can check that for any function $h$ defined on the 
appropriate set, we have
\begin{align}
    \sum_{t=1}^T h(i_t,\mathbf X_t,\mathbf C_t) &= 
    \sum_{i=1}^m \sum_{t:i_t=i} h(i_t,\mathbf X_t,
    \mathbf C_t)
     = \sum_{i=1}^m \sum_{\ell =1}^{n_i(T)} 
    h(i,\mathbf X_{i,\ell},\mathbf C_{i,\ell}). 
\end{align}
This yields
\begin{align}
    R_{T,1} &=\sum_{i\not\in I^*} n_i(T)\,\Delta_i\\
    R_{T,2} &= \sum_{i=1}^m \sum_{\ell=1}^{n_i(T)}  
        f^\circ_i (\mathbf X_{i,\ell},\mathbf{C}_{i,
        \ell}) = \sum_{i=1}^m \sum_{\ell=1}^{n_i(T)} 
        \big\{f^*_i (\mathbf C_{i,\ell}) - f_i(\mathbf 
        X_{i,\ell},\mathbf{C}_{i,\ell})\big\}\\
    R_{T,3} &= \sum_{i=1}^m \sum_{\ell=1}^{n_i(T)} 
        \big\{\mathbb E [f_{i}^\circ(\mathbf X_{i,\ell}, 
        \mathbf C_{i,\ell}) \mid \boldsymbol{f}, 
        \mathcal D^{i,\ell-1}] - f^\circ_{i} (\mathbf 
        X_{i,\ell}, \mathbf C_{i,\ell}) \big\}.
\end{align}
We will upper bound the three components of the 
regret, $R_{T,1}$, $R_{T,2}$ and $R_{T,3}$, separately.

\paragraph{Bounds on the arm selection regret $R_{T,1}$:} 
We have
\begin{align}
    R_{T,1}& \le \sum_{i\not\in I^*,n_i(T)>1} \sqrt{n_i(T)}\,
    \sqrt{n_i(T)}\,\Delta_i + \sum_{i=1}^m \Delta_i\\
    &\le \sum_{i\not\in I^*} \sqrt{n_i(T)} \big(
    \sqrt{\mathsf C_i\beta_i(n_i)\gamma_i(n_i)} + 2
    \rho_i(n_i)\big) + \sum_{i=1}^m \Delta_i\\
    &\le \bigg\{T\sum_{i\not\in I^*}\mathsf C_i 
    \beta_i(n_i)\gamma_i(n_i)\bigg\}^{1/2} 
    + \bigg\{4 T\sum_{i\not\in I^*} \rho_i^2(n_i)
    \bigg\}^{1/2} + \sum_{i=1}^m \Delta_i.
\end{align}
The first inequality above is obtained by simple 
algebra using the fact that $\Delta_i\ge 0$ for 
every $i$, the second inequality follows from 
\Cref{lem:sub_opt_play}, while the third 
inequality is a consequence of the Cauchy-Schwarz
inequality and the fact that the terms $n_i(T)$ 
sum to $T$. Since all the functions $\beta_i 
(\cdot)$ and $\rho_i(\cdot)$ are nondecreasing, 
we get
\begin{align}
    R_{T,1}& \le \sqrt{mT}\big(\sqrt{\mathsf C 
    \beta(T) \bar\gamma(T)} + 2 \rho(T)\big) + 
    \sum_{i=1}^m \Delta_i \le 2\bar c(T) \sqrt{mT} 
    + \|\boldsymbol{\Delta}\|_1. \label{eq:RT1a}
\end{align}
An alternative bound can be obtained by using 
\Cref{lem:sub_opt_play} to infer that
\begin{align}
    R_{T,1} &\le \sum_{i\not\in I^*} \bigg( \Big\{ 
    \frac{\sqrt{\mathsf C_i\beta_i(n_i)\gamma_i(n_i)} 
    + 2\rho_i(n_i)}{\Delta_i}\Big\}^2 + 1\bigg)\Delta_i\\
    &\le \sum_{i\not\in I^*} \bigg\{ 
    \frac{4c^2(n_i)}{\Delta_i} + \Delta_i\bigg\}
    \le 4c^2(T)\sum_{i\not\in I^*}  \frac{1}{\Delta_i} 
    + \|\boldsymbol{\Delta}\|_1.
    \label{eq:RT1b}
\end{align}

\paragraph{Bounds on the action selection regret $R_{T,2}$:} 
Let us use the shorthand notation $\lambda = 
{\varphi}\sqrt{\log(4md\psi/\delta)}$. 
In view of \Cref{lem:event}, on the event 
$\mathcal A(\delta) \subset \bigcap_{i=1}^m \mathcal A_i
(\delta/2m)$, for any arm $i\in [m]$ and for any $n\in
\mathbb N^*$, we have 
\begin{align}
    \sum_{\ell=1}^{n} \big(f_i^*(\mathbf C_{i,\ell}) 
    - f_{i}(\mathbf{X}_{i,\ell}, \mathbf{C}_{i,\ell})\big) 
    &\leq 
    2\sum_{\ell=1}^{n}\sqrt{\beta_{i}(\ell)}\, 
    \sigma_{i,\ell-1}(\mathbf{X}_{i,\ell}, \mathbf{C}_{i,\ell})
    + \frac{\lambda}{m}\\
    &\le  2\sqrt{\beta_{i}(n)}\sum_{\ell=1}^{n} 
    \sigma_{i,\ell-1}(\mathbf{X}_{i, \ell}, \mathbf{C}_{i,\ell}) + \frac{\lambda}{m}\\
    &\le  2\sqrt{n\beta_{i}(n)}\bigg\{\sum_{\ell=1}^{n} 
    \sigma_{i,\ell-1}(\mathbf{X}_{i, \ell}, \mathbf{C}_{i,\ell})
    \bigg\}^{1/2} + \frac{\lambda}{m}\\
    &\leq \sqrt{n\mathsf C_i\beta_{i}(n)
    \gamma_{i}(n)} + \frac{\lambda}{m}.\label{eq:69}
\end{align}
Here, for the first inequality, we used \eqref{eq:16}
and \eqref{eq:17}, for the second inequality, we used the 
fact that $\beta_i(\cdot)$ is a nondecreasing function, 
the third line follows from 
the Cauchy-Schwarz inequality, whereas the forth line 
is a consequence of inequality \eqref{eq:66}. This leads
to a first bound on the second component of the regret:
\begin{align}
    R_{T,2} & \le \sum_{i = 1}^m \sqrt{n_i\mathsf C_i\beta_{i}(n_i) \gamma_{i}(n_i)} + \lambda  
    \\
    &\le \sqrt{\mathsf C\beta(T)} \sum_{i = 1}^m 
    \sqrt{n_i\gamma_i(n_i)} + \rho(T)\\
    &\le \sqrt{mT\mathsf C\beta(T) 
    \bar\gamma(T)} + \rho(T) = \sqrt{mT}\,\bar c(T).
    \label{eq:RT2a}
\end{align}
An alternative bound can be obtained by 
replacing in \eqref{eq:69} $n$ by $n_i = n_i(T)$ and,
for $i\not\in I^*$, using \Cref{lem:sub_opt_play}:
\begin{align}
    \sum_{\ell=1}^{n_i} \big(f_i^*(\mathbf C_{i,\ell}) 
    - f_{i}(\mathbf{X}_{i,\ell}, \mathbf{C}_{i,\ell})
    \big) \le 2\sqrt{\kappa\beta(1)} + \frac{4 c^2(T)}{
    \Delta_i} .
\end{align}
The first term in the right hand side above comes from
the fact that if $n_i=1$, then instead of using 
\Cref{lem:sub_opt_play}, we can simply use the fact that
$\mathsf C_i\gamma_i(1) \le 4\kappa_i$. 
Summing over all $i\in[m]$, we get
\begin{align}
    R_{T,2} &\le \sqrt{T|I^*|}\,c(T) + 2m\sqrt{\kappa 
    \beta(1)} + 5c^2(T) \sum_{i\not\in I^*} 
    \frac1{\Delta_i}.\label{eq:RT2b}
\end{align}
\paragraph{Bounds on the stochastic-error-term $R_{T,3}$} 
Let us denote by $\mathcal F_{i,n}$ the $\sigma$-algebra
generated by $\{\mathcal D^{i,n},\boldsymbol{f})$ and 
define $M_{i,0} = 0$ and
\begin{align}
    M_{i,n} = \sum_{\ell=1}^n \big\{\mathbb E [f_{i}^\circ
    (\mathbf X_{i,\ell}, \mathbf C_{i,\ell}) \mid 
    \boldsymbol{f}, \mathcal D^{i,\ell-1}] - 
    f^\circ_{i} (\mathbf X_{i,\ell}, \mathbf C_{i,
    \ell})\big\},\qquad \text{for } n\in\mathbb N^*.
\end{align}
It is clear that $M_{i,n}$ is a $\mathcal F_{i,n} 
$-martingale, in the sense that $\mathbb E[M_{i,n}
\mid \mathcal F_{i,n-1}] = M_{i,n-1}$. To upper 
bound $M_{i,n}$, we will apply the Azuma-Hoeffding 
inequality \citep{Azuma,Hoeffding}. To this end, 
note that $f_i^\circ(\mathbf x,\mathbf c) \ge 0$ 
for every $\mathbf x$ and $\mathbf c$. Furthermore, 
as seen in the proof of \Cref{lem:Hoeffding}, the 
set of functions $f_i$ satisfying the inequalities
\begin{align}
    f_i^\circ(\mathbf X_{i,\ell},\mathbf C_{i,\ell}) &= 
    \max_{\mathbf x} f_i(\mathbf x,\mathbf C_{i,\ell})
    - f_i(\mathbf X_{i,\ell},\mathbf C_{i,\ell})\\
    &\le \Delta f_i^* \le d'r'\varphi\sqrt{\log(4m\smash{\bar d}
    \psi/\delta)}:=\lambda'
\end{align}
is of probability at least $1-\delta/2m$. This implies 
that conditionally to $\boldsymbol{f}$, $|M_{i,n} - 
M_{i,n-1}| \le \lambda'$. Therefore, we can apply the 
Azuma-Hoeffding inequality to infer that
\begin{align}
    \mathbb P\big( M_{i,n} \le \lambda'\sqrt{2n\log(1/\delta_1)}
    \mid \boldsymbol{f}\big)\ge 1- \delta_1 
\end{align}
with probability at least $1-\delta/2$ over the randomness
in $\boldsymbol{f}$. Choosing $\delta_1 = \delta/(2m\pi_n)$, 
and using the union bound, we arrive at
\begin{align}
    \mathbb P\big( M_{i,n} \le \lambda'\sqrt{2n\log( 2m
    \pi_n/\delta)},\ \text{for all }n\in\mathbb N^*, \ 
    i\in[m]\, \mid  \boldsymbol{f}\big)\ge 1- \frac{\delta
    }{2} 
\end{align}
with probability at least $1-\delta/2$ over the randomness
in $\boldsymbol{f}$. From this, we readily get
\begin{align}
    \mathbb P\big( M_{i,n} \le \lambda'\sqrt{2n\log(2m\pi_n/
    \delta)},\ \text{for all }n\in\mathbb N^*,\;i\in[m] 
    \big)\ge 1- {\delta}. 
\end{align}
Hence, on an event $\mathcal B(\delta)$ of probability 
at least $1-\delta$, we have 
\begin{align}
    R_{T,3} &= \sum_{i=1}^m M_{i,n_i(T)} \le 
    \lambda'\sqrt{2\log(2m\pi_T/\delta)} \sum_{i=1}^m 
    \sqrt{n_i(T)}\label{eq:RT3_}\\
    &\le \lambda'\sqrt{2mT\log(2m\pi_T/\delta)}. 
    \label{eq:RT3a}
\end{align}
The alternative upper bound on the term $R_{T,3}$
is obtained by combining \eqref{eq:RT3_} with 
\Cref{lem:sub_opt_play}. This yields
\begin{align}
    R_{T,3} &\le \rho(T) \bigg(\sqrt{T|I^*|} + m +
    2c(T)\sum_{i\not\in I^*} \frac{1}{\Delta_i} \bigg). 
    \label{eq:RT3b}
\end{align}

Combining upper bounds \eqref{eq:RT1a}, \eqref{eq:RT2a}
and \eqref{eq:RT3a}, we get
\begin{align}
    R_T &\le 3\sqrt{mT}\,\bar c(T) + \|\boldsymbol{
    \Delta}\|_1  
    + \lambda'\sqrt{2mT\log(2m\pi_T/\delta)}\\
    &\le 5\sqrt{mT}\,\bar c(T) + \|\boldsymbol{
    \Delta}\|_1  
\end{align}
with probability at least $1-2\delta$. 

Similarly, combining bounds \eqref{eq:RT1b}, \eqref{eq:RT2b}
and \eqref{eq:RT3b}, we get
\begin{align}
    R_T &\le 4c^2(T)\sum_{i\not\in I^*}  
    \frac{1}{\Delta_i} + \|\boldsymbol{\Delta}\|_1 + 
    \sqrt{T|I^*|}\,c(T) + 2m\sqrt{\kappa\beta(1)} + 
    5c^2(T)\sum_{i\not\in I^*} \frac1{\Delta_i}\\
    &\qquad + \rho(T) \Big(\sqrt{T|I^*|} + m + 
    2c(T)\sum_{i\not\in I^*}\frac{1}{\Delta_i}\Big).
\end{align}
Regrouping the terms, we arrive at
\begin{align}
    R_T &\le 11c^2(T)\sum_{i\not\in I^*}  
    \frac{1}{\Delta_i} + \|\boldsymbol{\Delta}\|_1 + 
    2\sqrt{T|I^*|}\,c(T) + m\big(2\sqrt{\kappa\beta(1)} + 
    \rho(T)\big).
\end{align}
This completes the proof of the proposition.
\end{proof}


\subsection{Proof of \Cref{th:1}}

We denote $\bar\varphi=\varphi\vee1$ and $\bar \psi_=\psi\vee 1$, and take $\pi_n=\pi^2n^2/6$. \Cref{prop:reg_bound} assumes that 
    \begin{align}
            \beta_{i}(n) &\geq  2 \log(4/\delta) + 4d_i
        \log(mr_id_i\pi_n),
    \end{align}
where the right-hand side can be upper bounded by
\begin{align}
    2 \log(4/\delta) + 4d_i
        \log(mr_id_i\pi_n)&\leq 4d_i\log(2\pi_nr_imd_i/\delta)\\
        &\leq8d_i\log\left(\frac{\pi\sqrt{r_i}}{3}\cdot\frac{md_in}{\delta}\right),
\end{align}
since $d_i\geq1$.

Now we analyse the assumption on $\rho_i(n)$:
\begin{align}
    \rho_i(n)\geq{\varphi}\sqrt{
        log(4m\smash{\bar d_i}\psi /\delta)} +  \dprimei r_i'\varphi 
        \log\big(4m(\dprimei \psi+2\pi_n)/\delta\big),
\end{align}
where the right-hand side can be upper bounded by
\begin{align}
    {\varphi}\sqrt{
        \log(4m\smash{\bar d_i}\psi /\delta)}& +  \dprimei r_i'\varphi 
        \log\big(4m(\dprimei \psi+2\pi_n)/\delta\big)\\
        &\leq\varphi\left(\log(4m\bar d_i\bar\psi/\delta)+\dprimei r_i'
        \log\big(4m(\smash{\bar d_i} \psi+2\pi_n)/\delta\big)\right)\\
        &\leq 2\varphi\dprimei (r_i'\vee 1)\log\left(4m\left(\frac{\smash{\bar d_i}\bar\psi}{\delta}+\frac{\pi^2n^2}{3}\right)\right)\\
        &\leq 2\varphi\dprimei (r_i'\vee 1)\log\left(8m\smash{\bar d_i}\bar\psi\pi^2n^2/(3\delta)\right), \ \text{as } a+b\leq2ab, \  \forall{a,b\geq 1}\\
        &\leq 4\varphi\dprimei(r_i'\vee 1)\log\left(\frac{2\sqrt{2}\pi\bar\psi}{\sqrt{3}}\cdot \frac{mn\smash{\bar d_i}}{\delta}\right).
\end{align}

Hence, $\beta_i(n)$ and $\rho_i(n)$ satisfy the conditions of \Cref{prop:reg_bound} if 
\begin{align}
    \beta_i(n)\wedge \rho_i(n)\geq 8\bar\varphi\smash{\bar d_i}(\bar r_i\vee 1)\log\left(5.14\bar\psi\sqrt{\bar r_i\vee 1}\cdot mn\smash{\bar d_i}/\delta\right).
\end{align}

By taking $\mathsf A_1=8\bar\varphi\bar\psi(\bar r_i\vee 1)$, then $\beta_i(n)$ and $\rho_i(n)$ satisfy the conditions of \Cref{th:1} whenever they satisfy the conditions of \Cref{prop:reg_bound}.

Under \Cref{prop} the regret is upper bounded
\begin{align}
    R_T &\leq 5\sqrt{mT}\,\big(\sqrt{\mathsf C
        \beta(T)\bar \gamma(T)} + \rho(T)\big) + 
        \|\boldsymbol{\Delta}\|_1\\
        &\leq 5\sqrt{\mathsf C\vee  1}\{\sqrt{mT}\big(\sqrt{
        \beta(T)\bar \gamma(T)} + \rho(T)\big)  
       \}+ \|\boldsymbol{\Delta}\|_1,
\end{align}
which implies $\mathsf A_2=5\sqrt{\mathsf C\vee  1}$.

For the instance dependent upper bound, \Cref{prop} implies that
\begin{align}
    R_T &\le 2\sqrt{|I^*| T}\,\big(\sqrt{\mathsf 
        C \beta(T)\bar \gamma(T)} + \rho(T)\big) + 
        22\big(\mathsf C\beta(T) \gamma(T) + 
        \rho^2(T)\big)\sum_{i\not\in I^*} \frac{1}{
        \Delta_i}\\ 
        &\qquad + \|\boldsymbol{\Delta}\|_1 + 
        2m\big(\sqrt{\kappa\beta(1)} + \rho(T)\big)\\
        &\leq 2\sqrt{\mathsf 
        C\vee1}\{\sqrt{|I^*| T}\,\big(\sqrt{ 
        \beta(T)\bar \gamma(T)} + \rho(T)\big)\}+22(\mathsf  C\vee 1)\big(\beta(T) \gamma(T) + 
        \rho^2(T)\big)\sum_{i\not\in I^*} \frac{1}{
        \Delta_i}\\
        &\qquad  + 
        2(\sqrt{\kappa}\vee 1)m\big(\sqrt{\beta(1)} + \rho(T)\big)+ \|\boldsymbol{\Delta}\|_1.
\end{align}
 Consequently, we take $\mathsf A_3=2(11\mathsf C\vee\sqrt{\kappa}\vee11)$.

In summary, if we take $\mathsf A_1=8\bar\varphi\bar\psi(\bar r_i\vee 1)$, $\mathsf A_2=5\sqrt{\mathsf C\vee  1}$, and  $\mathsf A_3=2(11\mathsf C\vee\sqrt{\kappa}\vee11)$ then \Cref{prop} implies \cref{th:1}.
 
\section{Additional theoretical results}

In this section, we provide some additional mathematical
results that support the discussion conducted after 
\Cref{th:1}.

\subsection{Bounds on the information gains}

\begin{table}[t]
    \centering
    \begin{tabular}{l|c|p{5.3cm}}
        \toprule
        Kernel & $\gamma_T$ & $\max_{\substack{n_i\in\mathbb{N}: \sum n_i=T}}\sum_{i=1}^m\gamma_{n_i}$, \\
        \midrule
        Matérn-$\nu$ \tablefootnote{\label{note1}Please see \cite{inf_gain_bounds}} & $\mathcal{O}(T^{\frac{d}{2\nu+d}}\log^{\frac{2\nu}{2\nu+d}}(T))$ & $\mathcal{O}(m^{\frac{2\nu}{2\nu+d}}T^{\frac{d}{2\nu+d}}\log^{\frac{2\nu}{2\nu+d}}(T)+m)$\\[3pt]
        Squared exponential \footnoteref{note1} & $\mathcal{O}(\log^{d+1}(T))$ & $\mathcal{O}(m\log^{d+1}(T))$\\[3pt]
        Cauchy spectral mixture \tablefootnote{Please see \cite{mixture_kernels}} & $\mathcal{O}(T^{\frac{2d^2+2d+1}{2d^2+2d+2}}\log(T))$ & $\mathcal{O}(m^{\frac{1}{2d^2+2d+2}}T^{\frac{2d^2+2d+1}{2d^2+2d+2}}\log(T)+m)$\\[3pt]
       $D$-dimensional feature-map \tablefootnote{Please see \cite{srinvas_kauss}} & $\mathcal{O}(D\log(T))$ & $\mathcal{O}(mD\log(T))$\\[3pt]
        $\beta$-polynomial eigendecay \footnoteref{note1}& $\mathcal{O}(T^{1/\beta}\log^{1-1/\beta}(T))$ & $\mathcal{O}(m^{1-1/\beta}T^{1/\beta}\log^{1-1/\beta}(T)+m)$\\[3pt]
    $\beta$-exponential eigendecay \footnoteref{note1}& $\mathcal{O}(\log^{1+1/\beta}(T))$ & $\mathcal{O}(m\log^{1+1/\beta}(T))$\\[3pt]
        \bottomrule
    \end{tabular}
    \caption{Bounds on maximum information gains and their 
    sums on a domain of dimension $d$.}
    \label{tab:sum_gamma}
\end{table}

This subsection aims to upper bound sums of the 
following form
\begin{align}\label{eq:target_sum_gamma}
    \max_{\substack{n_i\in\mathbb{N^+} \\ \sum n_i=T}}\sum_{i=1}^m\gamma_i(n_i),
\end{align}
where $\gamma_i$ are the maximum information gains 
for possibly different Gaussian processes (\textit{i.e.}, 
with different kernels) but satisfying our assumptions 
(0 mean, covariance function is bounded by $\kappa$, 
etc.). We set $\gamma_i(t)=0,\ \forall t\leq 0$ for 
convenience.

To upper bound \eqref{eq:target_sum_gamma}, we use 
results of \cite{inf_gain_bounds} that imply that 
\begin{itemize}
    \item for kernels with exponential eigendecay
    $\gamma_T \leq \mathcal{O}(\log^{1+\frac{1}{\beta_e}})$ 
    where $\beta_e$ is a positive constant,
    \item for kernels with polynomial eigendecay $\gamma_T 
    \leq \mathcal{O}(T^{\frac{1}{\beta_p}}\log^{1 - 
    \frac{1}{\beta_p}}T)$, where $\beta_p>1$.
\end{itemize}
As an example, for widely used kernels, such  Matern-$\nu$ 
$\gamma_T\leq\mathcal{O}(T^{\frac{d}{2\nu+d}}\log^{
\frac{2\nu}{2\nu+d}}T)$  and $\gamma_T\leq\mathcal{O}(
\log^{d+1}T)$ for the squared exponential kernel.



Hence, we write that $\gamma_i(T)\leq A_iT^{a_i}\log^{b_i}T+B_i$ for some $A_i, B_i, b_i>0$ and $0\leq a_i<1$. For simplicity, we take $A,B,b$ and $a$ to be the maximums of the corresponding constants, e.g. $a:=\max_{i\in[m]} a_i$. So we have that 
\begin{align}\label{eq:cond_gamma}
    \gamma_{i}(T)\leq A {T^{a}\log^{b}T} + B, \quad\forall 
    i \in[m].
\end{align}
We note that $B\geq\gamma_i(1),\ \forall i\in[m]$ which 
under the assumption that kernels are bounded by $\kappa$ 
is at most $\frac{1}{2}\log(1+\sigma^{-2}\kappa)$. So 
$B\geq\frac{1}{2}\log(1+\sigma^{-2}\kappa)$.

\begin{proposition}\label{pr:sum_of_gammas_bound_th}
    If the array $(\gamma_i(n))_{i\in[m],n\in\mathbb N^*}$
    satisfies \eqref{eq:cond_gamma}, then
    \begin{align}
        \max_{\substack{n_i\in\mathbb{N^+} \\ \sum n_i=T}}\sum_{i=1}^m\gamma_i(n_i)\leq A m^{1-a} 
        T^a \log^b {T} + mB.
    \end{align}
\end{proposition}

\begin{proof} Using that $x\mapsto \log^b x$ is increasing and 
that $x\mapsto x^a$ is concave, we upper bound \eqref{eq:target_sum_gamma} by
\begin{align}
     \max_{\substack{n_i\in\mathbb{N^+} \\ \sum n_i=T}}\sum_{i=1}^m\gamma_{i,n_i}&\leq A\max_{\substack{n_i\in\mathbb{R},\ n_i\geq1\\ \sum n_i=T}} \log^b T\sum_{i=1}^m n_i^a + m B\\
     &\leq Am\log^b T \max_{\substack{n_i\in\mathbb{R},\ n_i\geq1\\ \sum n_i=T}} \bigg(\frac1m \sum_{i=1}^m n_i\bigg)^a + m B\\
     & = Am^{1-a}T^a \log^b T  + m B
\end{align}
and the claim of the proposition follows.
\end{proof}

We illustrate the implications of 
\Cref{pr:sum_of_gammas_bound_th} for several widely 
used kernels in \Cref{tab:sum_gamma}. Our results 
also extend to composite kernels, i.e.\ sums and 
products of kernels. In particular, Theorem 3 of \citep{srin_kauss_cont} gives an upper bound on 
the maximum information gain of a sum kernel in 
terms of the information gains of its components. 
Similarly, Theorem 2 of \citep{srin_kauss_cont} 
shows that when the component kernels are finite‐rank, 
the information gain of their product can be bounded 
likewise by those of the components.

\subsection{On the uniqueness of the optimal arm}
\label{sec:arm_uniqueness}

This section motivates the uniqueness of the optimal 
arm in the scenario where $(f_i)_i$ are sampled from 
Gaussian processes.
We recall  that the reward of arm $i$ is 
\begin{align}\label{eq:app_c5_1}
    \mu_i=\mathbb{E}\left[\sup_{\mathbf{x\in\mathfrak{X}_{\mathbf X}}} f_i(\mathbf{x, C})\mid f_i\right].
\end{align}

We next show when $\mu_i\neq\mu_j$ for distinct $i$ and $j$.

\begin{lemma}\label{lem:c_2_5}
    Let $i\neq j$. If at least one of $\mu_i$ and $\mu_j$ is absolutely continuous, then $P(\mu_i=\mu_j)=0$.
\end{lemma}
\begin{proof}
    Note that $\mu_i$ is $\sigma(f_i)$  measurable  and $\mu_j$ is $\sigma(f_j)$ measurable. By \Cref{ass:2}, the random variables $f_i$ and $f_j$ are independent, which implies that  any random variables defined on their respective $\sigma$-algebras are also independent, in particular $\mu_i$ and $\mu_j$ are independent.

    We assume without loss of generality that $\mu_i$ is absolutely continuous. Under the established independence we write that
    \begin{align}
        P(\mu_i=\mu_j)&=(P_{\mu_i}\otimes P_{\mu_j})(\mu_i=\mu_j)=\int P_{\mu_i}(\mu_i=x) \ P_{\mu_j}(\rmd x)\\
        &=\int0 \ P_{\mu_j}(\rmd x)=0,
    \end{align}
    which concludes the proof.
\end{proof}

It follows from \Cref{lem:c_2_5} that if all $\mu_i$, except possibly one, are absolutely continuous, then almost surely no two distinct ones are equal.

The next lemma provides sufficient conditions for $\mu_i$ to be absolutely continuous.

\begin{lemma}\label{lem:c_2_lem_6}
    Let $|\mathfrak{X}_{\mathbf{C}(i)}|<\infty$ and $k_i$ be strictly positive definite, then $\mu_i$ is absolutely continuous. 
\end{lemma}
\begin{proof}

Our \Cref{ass:2} implies that $f_i$ is 
almost surely continuous and hence, thanks to 
the compactness assumption made in \Cref{ass:1}, 
attain their maxima on $\mathfrak{X}_{\mathbf{X} 
(i)}$. We may rewrite the expectation in 
\eqref{eq:app_c5_1} as a sum, assuming that
$\mathfrak{X}_{\mathbf{C}(i)} = \{\mathbf c_1,
\ldots,\mathbf c_J\}$:
\begin{align}
    \mu_i& = \sum_{\mathbf{c\in\mathfrak{X}_{
    \mathbf{C}(i)}}} P(\mathbf{C}=\mathbf{c}) 
    f_i\left(\mathbf{x}^*_{i}(\mathbf{c}), 
    \mathbf{c}\right)\\
    & = \max_{\mathbf x_1,\ldots,\mathbf x_J}
    \sum_{j=1}^J \nu_j f_i(\mathbf{x}_j, 
    \mathbf{c}_j).
\end{align}
It is clear that the process 
\begin{align}
    F_i(\mathbf{x}_1,\ldots,\mathbf{x}_J ) = 
    \sum_{j=1}^J \nu_k f_i(\mathbf{x}_j, 
    \mathbf{c}_j)
\end{align}
is centered and Gaussian. For $\vec{\mathbf{x}} 
= (\mathbf{x}_1,\ldots,\mathbf{x}_K )$ and 
$\vec{\mathbf{x}}' = (\mathbf{x}'_1,
\ldots,\mathbf{x}'_K )$, we have
\begin{align}\label{eq:big_f_cov}
    \mathbb E[F(\vec{\mathbf{x}})F(\vec{
    \mathbf{x}}')] = \boldsymbol{\nu}^\intercal
    \mathbf K_i(\vec{\mathbf{x}},\vec{\mathbf{x}}')
    \boldsymbol{\nu},
\end{align}

where $\boldsymbol{\nu}^\intercal=(\nu_1,\dots, \nu_J)$ and
\begin{align}
   \mathbf K_i(\vec{\mathbf{x}},\vec{\mathbf{x}}')=\begin{pmatrix}
k_i((\mathbf{x}_1,\mathbf{c}_1), (\mathbf{x}'_1,\mathbf{c}_1)) & \dots & k_i((\mathbf{x}_1,\mathbf{c}_1), (\mathbf{x}'_J,\mathbf{c}_J))\\
\vdots & \ddots & \vdots \\
k_i((\mathbf{x}_J,\mathbf{c}_J), (\mathbf{x}'_1,\mathbf{c}_1)) & \dots & k_i((\mathbf{x}_J,\mathbf{c}_J), (\mathbf{x}'_J,\mathbf{c}_J))
\end{pmatrix}.
\end{align}

Note that $\mathbf{K}_i(\vec{\mathbf{x}}, \vec{\mathbf{x}}')$ is strictly positive definite since $k_i$ is strictly positive definite. So \eqref{eq:big_f_cov} is strictly greater than zero for all $\mathbf{x}$ and $\mathbf{x}'$ from their respective domains.

We have that $F_i$ is a zero-mean Gaussian process whose index set is compact and separable, as it is a compact subset of Euclidean space, and its variance is zero nowhere. This implies that $\sup_{\mathbf{x}}F(\vec{\mathbf{x}})$
is absolutely continuous \citep[Corollary, page 2]{Lifshits1984}, and so is $\mu_i$.
\end{proof}

Consequently, if we have that for all $i$, except possibly one, $|\mathfrak{X}_{\mathbf{C}(i)}|<\infty$ and $k_i$ are strictly positive definite, we can invoke \Cref{lem:c_2_lem_6} and \Cref{lem:c_2_5} to conclude that  all $\mu_i$ are almost surely distinct. So $I^*$ set contains only one element almost surely.

\end{document}